\documentclass{article}
\usepackage{amsmath,amsfonts,amsthm,amssymb}
\allowdisplaybreaks
\usepackage{graphicx}
\usepackage{url}
\usepackage[dvipsnames,svgnames,table]{xcolor}
\usepackage{algorithm}
\usepackage{bm}
\usepackage{stackrel}
\usepackage{enumitem}
\usepackage[parfill]{parskip}
\usepackage{tabularx}
\usepackage{bbm}
\usepackage{caption}
\usepackage{subcaption}  %
\usepackage{placeins}
\usepackage{float}
\usepackage{afterpage}
\usepackage{textcomp}

\usepackage[normalem]{ulem}
\useunder{\uline}{\ul}{}
\usepackage{natbib}
\usepackage{pythonhighlight}
\usepackage{textcomp} %
\usepackage{makecell} %
\usepackage{todonotes} %
\usepackage{hyperref} %
\usepackage{microtype}
\usepackage{graphicx}
\usepackage{booktabs}

\usepackage{amsmath}
\usepackage{amssymb}
\usepackage{mathtools}
\usepackage{amsthm}
\usepackage[capitalize,noabbrev]{cleveref}
\usepackage{tcolorbox}
\usepackage{tabularray}
\usepackage{float}
\usepackage{placeins}
\usepackage{perpage}
\usepackage{lipsum}
\theoremstyle{plain}

\hypersetup{
    hidelinks
}

\newcommand{\framework}{{\it CheapVS}}
\newcommand\blfootnote[1]{%
  \begingroup
  \renewcommand\thefootnote{}\footnote{#1}%
  \addtocounter{footnote}{-1}%
  \endgroup
}

\newcommand{\Dc}{{\mathcal{D}}}

\newcommand{\Lc}{{\mathcal{L}}}
\newcommand{\Mc}{{\mathcal{M}}}

\newcommand{\Ebb}{{\mathbb{E}}}

\newcommand{\Rbb}{{\mathbb{R}}}

\DeclareMathOperator*{\argmax}{arg\,max}

\usepackage[accepted]{icml2025}
\usepackage{blindtext}
\usepackage{algorithm}
\usepackage{algorithmic}
\usepackage{amsmath, amssymb}

\icmltitlerunning{Preferential Multi-Objective Bayesian Optimization for Drug Discovery}

\begin{document}
\twocolumn[
\icmltitle{Preferential Multi-Objective Bayesian Optimization for Drug Discovery}

\icmlsetsymbol{equal}{*}

\begin{icmlauthorlist}
\icmlauthor{Tai Dang}{equal,rhf,stanford_cs,stanford_med}
\icmlauthor{Long-Hung Pham}{equal,imperial}
\icmlauthor{Sang T. Truong}{equal,stanford_cs}
\icmlauthor{Ari Glenn}{stanford_med}
\icmlauthor{Wendy Nguyen}{rhf}
\icmlauthor{Edward A. Pham}{stanford_med}
\icmlauthor{Jeffrey S. Glenn}{stanford_med}
\icmlauthor{Sanmi Koyejo}{stanford_cs}
\icmlauthor{Thang Luong}{rhf}
\end{icmlauthorlist}

\icmlaffiliation{rhf}{Rethink Healthcare Foundation - RHF.AI}
\icmlaffiliation{stanford_med}{Stanford Medicine}
\icmlaffiliation{imperial}{Chemistry Department, Imperial College London}
\icmlaffiliation{stanford_cs}{Stanford Computer Science}
\icmlcorrespondingauthor{Tai Dang}{taitdang@stanford.edu}
\icmlcorrespondingauthor{Long-Hung Pham}{l.pham23@imperial.ac.uk}
\icmlcorrespondingauthor{Sang T. Truong}{sttruong@cs.stanford.edu}
\icmlcorrespondingauthor{Thang Luong}{lmthang@stanford.edu}

\icmlkeywords{Machine Learning, ICML}

\vskip 0.3in
]

\printAffiliationsAndNotice{\icmlEqualContribution}
\begin{abstract}
    Despite decades of advancements in automated ligand screening, large-scale drug discovery remains resource-intensive and requires post-processing hit selection, a step where chemists manually select a few promising molecules based on their chemical intuition. This creates a major bottleneck in the virtual screening process for drug discovery, demanding experts to repeatedly balance complex trade-offs among drug properties across a vast pool of candidates. To improve the efficiency and reliability of this process, we propose a novel human-centered framework named \framework{} that allows chemists to guide the ligand selection process by providing preferences regarding the trade-offs between drug properties via pairwise comparison. Our framework combines preferential multi-objective Bayesian optimization with a docking model for measuring binding affinity to capture human chemical intuition for improving hit identification. Specifically, on a library of 100K chemical candidates targeting EGFR and DRD2, \framework{} outperforms state-of-the-art screening methods in identifying drugs within a limited computational budget. Notably, our method can recover up to 16/37 EGFR and 37/58 DRD2 known drugs while screening only 6\% of the library, showcasing its potential to significantly advance drug discovery.
\end{abstract}

\section{Introduction}
\begin{figure}[htb!]
\centering
\includegraphics[width=0.49\textwidth]{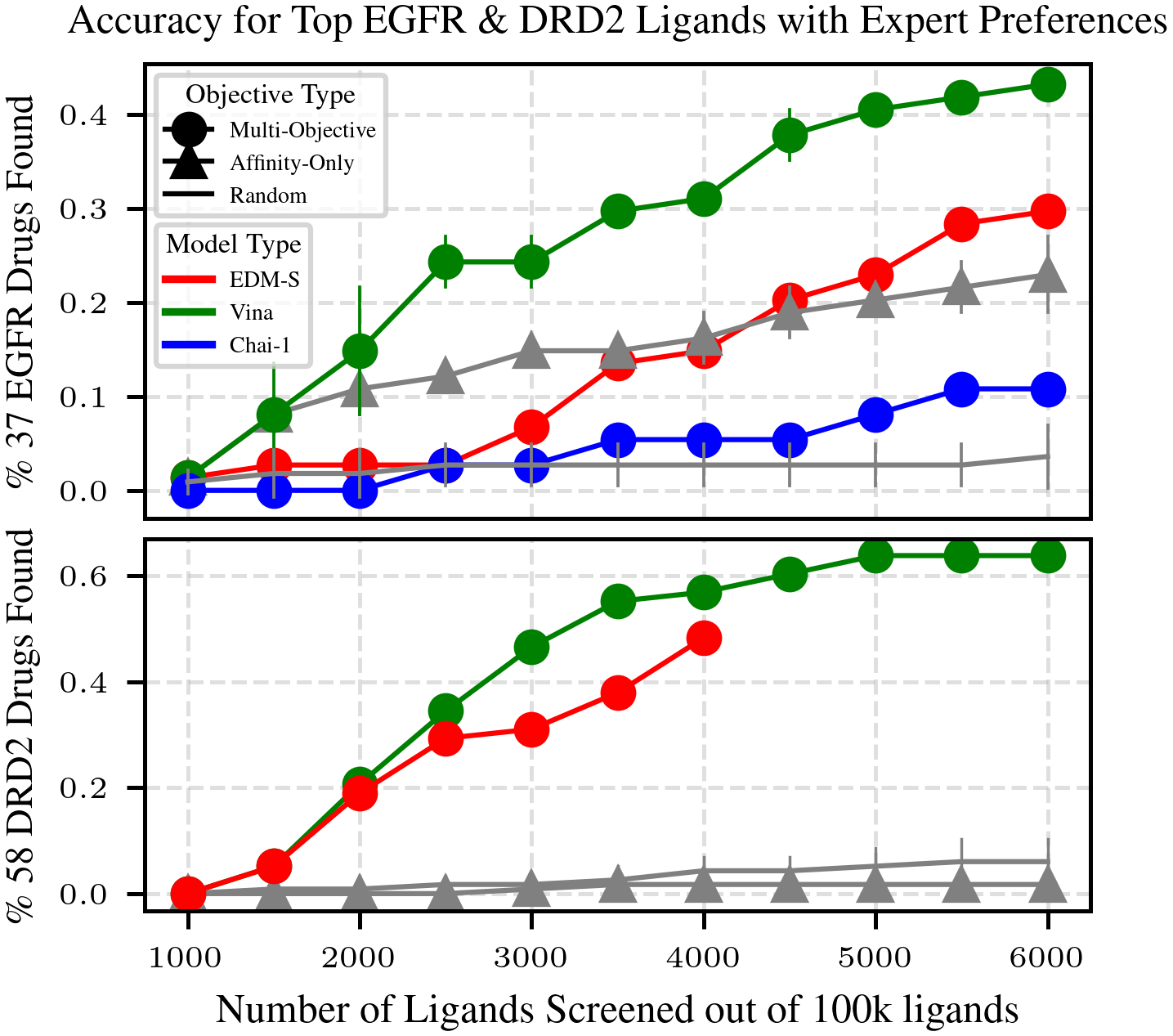}
\caption{Chemist-guided Active Preferential Virtual Screening performance in identifying EGFR and DRD2 drugs. The search is conducted on a 100K ligand library, screened for a maximum of 6\% of the library. The plot compares different methods for structure-based binding affinity measurement (Vina, EDM-S, Chai-1) and objective types. The y-axis shows the percentage of the top number of approved drugs identified, while the x-axis represents the number of ligands screened. Multi-objective optimization (circles) across all methods for affinity measures outperforms affinity-only selection (triangles) and random screening (gray line). Error bars indicate 1 standard deviation.}
\label{fig:moavs_hp}
\end{figure}

Virtual screening (VS) is a key pillar of modern computational drug discovery, acting as a rapid sift through massive molecular libraries-ranging from millions to billions of compounds-to identify a set of ``hits'' with promising therapeutic potential \cite{shoichet2004virtual, lyu2023expansionlib}. At the core of VS lies hit selection: the practical step in which a set of candidate compounds is manually chosen from top-ranked docking results, informed not only by binding affinity scores but also by key factors such as solubility, toxicity, and pharmacokinetic properties, all of which collectively determine a compound’s potential. Despite its centrality to drug discovery, VS and hit selection remain both resource-intensive. \blfootnote{Code and data are at
\href{https://github.com/vietai/cheapvs}{github.com/vietai/cheapvs}.}Traditional pipelines rely on exhaustive docking of the entire library, which demands substantial time and computational resources \cite{lyu2019ultra, gorgulla2020opensourcedrug}. Moreover, human expertise is required in the loop: medicinal chemists must examine the results to finalize which hits are worthy of costly experimental validation. In large-scale campaigns with millions of compounds, this process quickly becomes bottlenecked by both the computational costs and the limited bandwidth of experts. On top of the issue, vast computational resources are spent on characterizing unpromising (later-known low-scored) compounds, even though only a small fraction of top-ranked molecules typically move forward for hit selection and experimental validation. To address this problem, recent methods have combined active learning with binding affinity prediction to query compounds based on predicted binding affinity, substantially reducing computational overhead while maintaining high accuracy \cite{molpal, RosettaVS, gentile2020}.

Although binding molecules are good starting points for screening campaigns, the drug discovery process, in its entirety, is a complex multi-objective optimization problem. Indeed, VS presents a unique challenge due to its operation in a high-dimensional search space where these objectives (e.g., binding affinity, solubility, toxicity, pharmacokinetic properties, etc.) exhibit complex and often poorly understood interdependencies \cite{li2011balancing}. For instance, adding bulky functional groups can enhance binding affinity but simultaneously lower solubility or increase off-target toxicity, complicating the search for high-potential candidates. Balancing competing properties is key to robust drug leads. While single-objective active-learning approaches \cite{molpal, gentile2020} have shown promise in efficiently identifying top-scored molecules from large-scale libraries, the screening process still overlooks important considerations that medicinal chemists weigh in real-world pipelines, such as synthetic accessibility, stability, toxicity, etc. Thus, much computational power is still wasted on molecules with poor profiles other than binding affinity. This disconnection also highlights the critical role of domain expertise, balancing multiple factors that purely physics-based methods often fail to capture. Unfortunately, while invaluable, the expert-driven hit selection process is labor-intensive when scaled to large candidate pools.

To address these limitations, we present \framework{} (CHEmist-guided Active Preferential Virtual Screening) to assist chemists in expert-guided VS by leveraging a preferential multi-objective Bayesian optimization (BO) toolbox. By translating expert chemists’ nuanced understanding into multi-objective utility functions - incorporating factors such as binding affinity, solubility, or toxicity—our framework ensures that computational optimization captures subtle trade-offs that purely physics-based methods often overlook. This expert-guided approach refines the VS process, prioritizing candidates based on broader criteria crucial for downstream development. In doing so, we aim to make the hit identification process more efficient and aligned with expert preference and, ultimately, more effective in discovering promising drug leads from vast chemical spaces.

Preference ranking relies on the availability of a good measurement of ligand properties. An important measurement is the binding affinity between the ligand and the target protein. Recent breakthroughs such as AlphaFold3 \cite{alphafold3} and Chai-1 \cite{chai} have promised better measurement of binding affinity on a wide range of targeted proteins. Unfortunately, these methods are expensive, and the lack of understanding of their accuracy-efficiency trade-off has made it difficult for practitioners to select a suitable method for large scale VS. We compare the accuracy and efficiency of the physics-based and diffusion-based approaches, showing that while existing diffusion models are promising, their efficiency is far away from practical VS. Introducing a lightweight diffusion model, we significantly improve the efficiency of these tools while maintaining high performance, suggesting a path toward making deep learning models practical for VS.

In summary, our key contributions are:
\begin{itemize}[left=0pt]    
    \item \textbf{Eliciting Expert Preference for Efficient Virtual Screening:} We optimize trade-offs among interdependent drug properties by leveraging chemists' intuition through preference learning, translating domain knowledge into a latent utility function for more efficient VS.

    \item \textbf{Understanding Accuracy-Efficiency Trade-Off in Docking Models:} We evaluate the accuracy-efficiency trade-off of the physics-based and diffusion-based approaches and use data augmentation to significantly improve the efficiency of the diffusion docking model.

    \item \textbf{Efficient Multi-Objective Virtual Screening:} \framework{} considers various candidates' properties to simultaneously optimize them, such as binding affinity and toxicity, moving beyond single-objective paradigms.
\end{itemize}

\section{Related Work}
\paragraph{Efficient Decision Making in Virtual Screening} VS \cite{lionta2014structure, virtual-sreening} is a computational strategy for selecting promising molecules from large chemical libraries. Traditional high-throughput VS (HTVS) often employs computationally expensive structured-based binding affinity measurement methods \cite{gnina, smina, vina, lyu2019ultra}. While the effectiveness of ultra-large libraries is debated \cite{clark2020virtual}, their use in structure-based drug design has seen an increase in popularity \cite{gorgulla2020opensourcedrug, acharya2020supercomputerbasedensemble}. However, docking billions of compounds is computationally demanding \cite{gorgulla2020opensourcedrug}. Therefore, active learning strategies, such as MolPAL \cite{molpal}, improve efficiency by integrating optimal model-based sequential decision-making with docking. By training a machine learning model on initial binding affinities, MolPAL predicts binding affinities on the entire set and strategically selects subsequent compounds, significantly reducing the number of docking calculations while ensuring reliable performance.

\paragraph{Expert Preference in Virtual Screening} Multi-objective BO (MOBO) \cite{MOSBO} tackles the challenge of optimizing multiple, potentially conflicting objectives. A common approach uses the Expected Hypervolume Improvement (EHVI) \cite{eipareto, daulton2021parallelbayesianoptimizationmultiple}, while other strategies include Predictive Entropy Search, Max-value Entropy Search, and the Uncertainty-Aware Search Framework \cite{hernándezlobato2016predictiveentropysearchmultiobjective, belakaria2020maxvalueentropysearchmultiobjective, belakaria2022uncertaintyawaresearchframeworkmultiobjective}. ParEGO \cite{parEGO} addresses computationally expensive problems using landscape approximations. Recent work extends MOBO to high-dimensional spaces \cite{morbo}, accelerates VS, molecular optimization, and reaction optimization \cite{fromer2023paretooptimizationacceleratemultiobjective, zhu2023sampleefficientmultiobjectivemolecularoptimization, reaction-optimization}. However, many MOBO methods still lack mechanisms to effectively incorporate domain expert insights during the search process, which is a critical need in VS. \framework{} builds on this MOBO foundation \cite{MOSBO, pbo1, brochu2010tutorial} by introducing a preference learning framework that guides optimization towards solutions aligned with expert knowledge in VS. While prior work \cite{choung2023extracting} explores expert preferences via SMILES-based rankings, they do not consider ligand properties, which limits the optimization process

\paragraph{Measurement of Binding Affinity via Diffusion Model} Diffusion-based generative models have gained significant attention for their ability to model complex data distributions through iterative refinements of noisy inputs. Grounded on denoising score-matching \cite{hyvarinen2005estimation, song2019generative}, these models leverage a governing ordinary differential equation. The denoiser minimizes the mean squared error loss. Modern machine-learning models for structure-based binding affinity prediction often rely on experimentally verified structures from the Protein Data Bank (PDB)\cite{wwpdb2019protein}. Although the PDB offers thousands of structures, it contains only around 40k ligands. To broaden coverage, researchers often generate additional data, e.g., PDBScreen \cite{cao2024pdbScreen} introduces ``decoy'' ligands presumed not to bind the protein, while PigNet and CarsiDock \cite{pignet, carsidock} use techniques like re-docking, cross-docking, and random docking from large commercial libraries. These methods expand protein-ligand diversity, enabling models to improve their generalization for docking.

\section{Preliminary}
\label{sec:preliminary}
We briefly describe the VS setup. Given a target protein $\rho$ and a screening library $\Lc = \{ \ell_1, \ldots, \ell_N \}$, the goal is to select the top $k$ ligands with the highest potential to succeed as drugs. A value is assigned to each ligand toward this goal so they can be ranked. It is common in practice to use a molecular property vector of ligand $\ell$, denoted as $x_\ell$, as a proxy of drug likeliness. The vector can include various properties, such as binding affinity, toxicity, and solubility: $x_\ell = [x_\ell^\text{aff}, x_\ell^\text{toc}, x_\ell^\text{sol}]$. However, evaluating these properties can be costly and time-consuming for large molecular libraries. To address this, active screening methods are used to explore the chemical space efficiently. These methods start with a small, randomly chosen fraction of the ligand library as the initial training set. The properties of this small set of ligands, often binding affinity, are measured, which are then used to train a surrogate model. Once the initial model is established, the optimization proceeds iteratively. In each cycle, newly measured ligands update the model with their latest data. The updated surrogate model evaluates the remaining compounds, and an acquisition function $\alpha$ ranks them by balancing exploration and exploitation. The top-ranked candidates are selected, their properties are measured, and the resulting data is used to update the surrogate model further. This process continues until the termination criteria is reached. For evaluation, regret and top-k accuracy are the main metrics used in the screening procedure. \textit{Regret} at iteration $i$ is defined as $R_i = U^* - U(i)$, where $U^*$ is the highest possible utility in the library and $U(i)$ is the highest utility found at iteration $i$. Importantly, $U^*$ can only be determined if affinity computations are carried out for the entire library, making it a post-hoc evaluation metric rather than a run-time metric. Lower regret indicates a better screening strategy. \textit{Top-$k$ Accuracy} measures the proportion of correctly identified compounds within the \textit{top-$k$} set, where the top-$k$ corresponds to the compounds with the highest utility value. 

Measuring molecular properties is essential for eliciting expert preference. One of the most important properties is binding affinity, denoted as $x_{\ell, \rho}^{\text{aff}}$. Unfortunately, this objective is computationally expensive to estimate because it requires \textit{searching for an energetically optimal 3D structure of the ligand inside the protein binding pocket}:
\begin{equation}
    x_{\ell, \rho}^{\text{aff}} = \min_{\ell_{\text{3D}} \in \Rbb^{3 \times N_\ell}} h(\ell_{\text{3D}},\rho) \quad \forall \ell \in \Lc
    \label{binding-affinity-measure}
\end{equation}
where $h$ is the physics-based affinity scoring function based on the atomic interaction between the ligand and the target protein, $N_\ell$ is the number of atoms in ligand $\ell$ (which is typically several dozen), and $\ell_{\text{3D}}$ is the 3D coordinate vector of ligand $\ell$. Traditional docking methods use heuristics to search through the vast conformation space $\Rbb^{3 \times N_\ell}$, rendering an intractable procedure when applied at scale to all ligands in the library. Diffusion docking models have been introduced to bias the above optimization toward a 3D structure that geometrically fits the binding pocket. Given a protein target $\rho$, a ligand, and a corresponding experimentally obtained 3D binding pose, the training objective of diffusion docking model $p_\theta$ is to find the 3D pose that best fits the binding pocket in terms of mean squared error (MSE) loss. After training, the optimal 3D pose could be obtained rapidly without exhaustive search through sampling process $\ell_{\text{3D}} \sim p_\theta(\ell, \rho)$. The optimal pose is then used with the scoring function $h$ to measure binding affinity.

\section{Methods}
\begin{figure*}[t!]
\centering
    \includegraphics[width=0.83\textwidth]{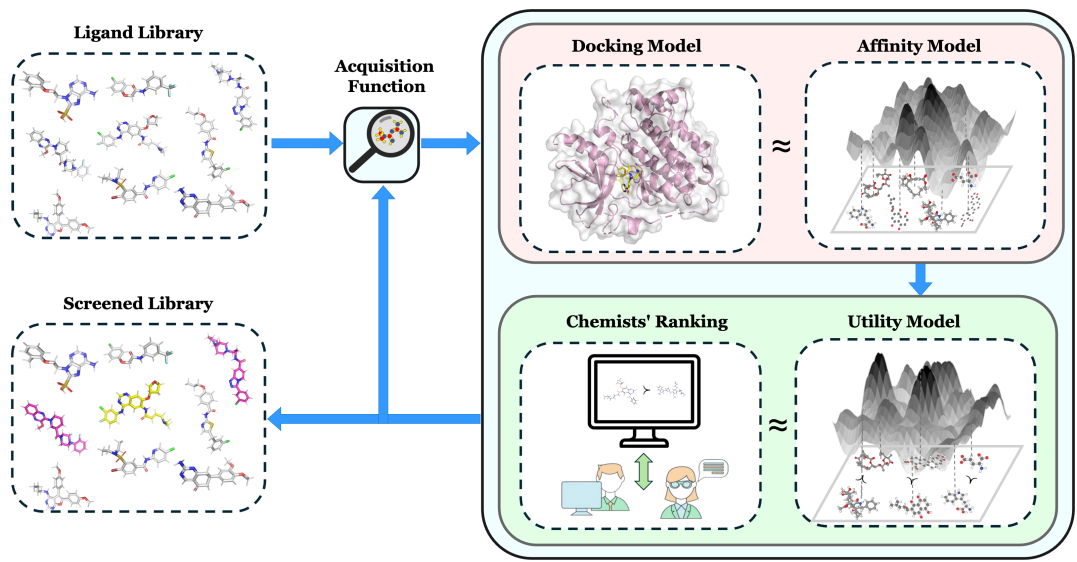}
    \caption{Overview of Chemist-guided Active Preferential Virtual Screening (\framework{}). Ligands from a large library are selected using an acquisition function and evaluated through structure-based affinity models. Chemists provide preference rankings, which inform a utility model to refine the selection process. The screened library iteratively improves, prioritizing ligands with desirable properties. Yellow-colored ligands represent found drug compounds, while purple ligands indicate screened compounds.}
\label{fig:cheapvs_overview}
\end{figure*}

We introduce a comprehensive methodology comprised of three components: preference modeling, active ligand selection, and ligand property measurement. An overview of the complete pipeline is presented in Figure~\ref{fig:cheapvs_overview}. 

\paragraph{Preference Modeling} Balancing multiple objectives, such as affinity, toxicity, and solubility, is necessary to identify drug candidates. This is challenging as the optimal trade-offs are unknown. We elicit expert preferences to guide the optimization process. These preferences are gathered through pairwise comparisons of ligand properties, where experts indicate which of the two ligands, $\ell_1$ and $\ell_2$, is more desirable based on multiple criteria. We assume the probability of preferring ligand $\ell_1$ over $\ell_2$ follows the Bradley-Terry model, generated via a utility function $f$ mapping from ligand property vector $x_\ell$ to a utility scalar:
\begin{equation}
    p(\ell_1 \succ \ell_2 \mid x_{\ell_1}, x_{\ell_2}) = \sigma(f(x_{\ell_1}) - f(x_{\ell_2}))
\end{equation}
where $\sigma$ is the logistic function. We model the distribution of the latent utility function with a Gaussian process (GP), assuming  $f \sim \mathcal{GP}(\mu(X), k(X, X'))$, where $\mu(X)$ and $k(X, X')$ denote the mean and kernel functions. We train $f$ using a dataset $\Dc_{f} = \{(x_{\ell_i}, x_{\ell_j}, y_{e_{ij}})\}$, where each pair $(x_{\ell_i}, x_{\ell_j})$ consists of ligand property vectors, and $y_{e_{ij}}$ is the expert preference label indicating whether $\ell_i$ is preferred over $\ell_j$. For posterior estimation, we employ the Laplace approximation. The choice of GPs for surrogate models is motivated by their superior performance over neural networks, as detailed in Appendix \ref{sec:surrogate}.

\paragraph{Measurement of Binding Affinity via Diffusion Model}
A large-scale diffusion docking model holds the promise to accelerate finding optimal binding poses in comparison to traditional searches. Unfortunately, sampling from the diffusion model is still an expensive process, especially for a large model, making it impractical to use diffusion for large-scale screening. Here, we aim to train a lightweight diffusion model that is highly computationally tractable for a large-scale library while aiming for a minimal reduction in binding affinity estimation. Toward this goal, we curate a large, diverse diffusion training dataset:
\begin{enumerate}
    \item From PDBScan22 \citep{flachsenberg2023redocking}, we remove non-drug-like ligands (e.g., solvents) to retain biologically relevant molecules (e.g., amino acids), ensuring the dataset focuses on drug-like ligands in realistic protein environments (see Appendix \ref{sec:appendix_preliminary_analysis}), yielding 180,000 high-quality protein-ligand pairs.
    \item To address low ligand diversity, we leverage the Papyrus dataset \cite{bequignon2023papyrus}, containing 260,000 active ligands across about 1,300 UniProt IDs. Molecules with reliable and good activity data (pChemBL $>$ 5) matching curated PDBScan22 structures were retained. Using Conforge \cite{seidel2023conforge}, we generate 50 conformers per molecule (RMSD $>$ 0.2 \AA). Pharmacophores from PDBScan22 are extracted with CDPKit \cite{CDPKit} and ligands are aligned based on shape and electrostatic properties. The aligned poses are minimized for binding affinity with Smina \cite{smina}.
\end{enumerate}
After training the diffusion on this curated data, for each ligand, we generate $128$ candidate conformation and greedily select the one with the lowest binding affinity to solve Equation \ref{binding-affinity-measure} by inference-time best-of-N search. One can train a diffusion policy to solve this equation directly via reinforcement learning, and we defer this to future work.

Even with a highly efficient diffusion model, measuring binding affinity on a vast ligand library remains intractable. We further tackle this problem by introducing a protein-specific surrogate model $g_\rho$ that directly predicts binding affinity for a given ligand. Here, we use GP to model the affinity distribution of $g_\rho$ with a Gaussian likelihood following \cite{brochu2010tutorial}. We featurize the ligand $\ell$ by its Morgan fingerprint $\ell_\Mc$, a fixed-length vector encoding its substructural features: $\hat{x}^{\text{aff}}_{\ell, \rho} = g_\rho(\ell_\Mc)$, where $g_\rho$ is supervised trained by leveraging data from diffusion model $\Dc_{g_\rho} = \{ (\ell_{\Mc, i}, h(\ell_{\text{3D}, i}) : \ell_{\text{3D}, i} \sim p_\theta(\ell_i, \rho))\}_i$
\paragraph{Active Virtual Screening with Expert Preference} We have introduced a latent utility function and a scalable method for measuring the binding affinity of a large ligand library, which can provide a measure for drug likeliness $f \circ g_\rho(\ell)$ of ligand $\ell$ that balances various objectives aligning with human preference. Since various approximations have been made in favor of computational tractability, one should not greedily optimize for $f \circ g_\rho$. The composition of utility and affinity functions allows uncertainty in $g_\rho$ to propagate through and impact decision quality if not handled carefully. Here, the optimal ligand for further measurement is the maximizer of the expected value of the acquisition function $\alpha$, marginalizing over posterior predictive distribution induced by $g_\rho$ for a given acquisition function $\alpha$ and a posterior distribution over $f$:
\begin{equation}
    \ell^* = \argmax_{\ell \in \Lc} \Ebb_{p(x_{\ell, \rho}^{\text{aff}} \mid \ell, \Dc_{g_\rho})} \alpha(p(f | \Dc_f), x_\ell).
\end{equation}
where the expectation is approximated using Monte Carlo sampling, and ligand pairs for expert preference queries are randomly chosen among the top-$k$ candidates with the highest acquisition values.  The full pseudocode for CheapVS's algorithm can be found in Appendix~\ref{sec:pseudocode}.

\section{Experiments}
\label{sec:experiments}

We conduct the experiments in three main stages. First, we investigate how well the utility model can learn from preference data, using both synthetic benchmark functions and real human-labeled preferences. Second, we explore the accuracy-speed tradeoff in affinity measurement by comparing our lightweight diffusion model (EDM-S) against Chai-1 and Vina, analyzing how computational efficiency impacts optimization performance. Finally, we integrate preference learning, molecular docking, and virtual screening into a comprehensive drug discovery pipeline, primarily targeting the EGFR and DRD2 proteins. For our main study, we utilize $\epsilon$-Greedy as the main acquisition function. A full analysis of additional strategies can be found in Appendix~\ref{sec:appendix_acq} and ~\ref{sec:synthetic}. Additionally, we compute physicochemical properties using \cite{rdkit} and incorporate ADMET predictions (absorption, distribution, metabolism, excretion, and toxicity) \cite{swanson2024admet} to account for realistic multi-objective trade-offs in drug discovery. Our study focuses on 3 Research Questions (RQ):

\begin{itemize}
    \item \textbf{RQ1}: How effectively can the latent utility function learn and approximate the underlying true utility from both synthetic and expert pairwise comparisons?
    \item \textbf{RQ2}: How does the accuracy-efficiency trade off in diffusion docking model impact the virtual screening process and what is a promising path to improve the efficiency of diffusion docking model? 
    \item \textbf{RQ3}: How effectively does \framework{} identify clinically relevant drug ligands using multi-objective optimization compared to affinity-only baseline?
\end{itemize}

\subsection{\mbox{Preference Elicitation from Pairwise Comparisons}}
\label{sec:utility_learning}

To answer \textit{RQ1}, we examine how well preference learning can correctly identify preferences using ligand properties (binding affinity, lipophilicity, molecular weight, and half-life) as input. For synthetic data, we generate 1,200 pairwise labels via functions--Ackley, Alpine1, Hartmann, Dropwave, Qeifail, and Levy. In contrast, for real human data, experts provide rankings on the EGFR target to form pairwise comparisons. All experiments are conducted under an 80/20 split and 20-fold cross-validation and evaluate model performance with accuracy and ROC AUC.

Table~\ref{tab:preference_results} indicates that our preference learning framework consistently achieves high accuracy and ROC AUC, demonstrating robust recovery of the latent utility function. While there is some variability across different synthetic functions, the overall trend confirms strong performance. Similarly, preliminary results on real human data show competitive performance, with an accuracy of approximately 80\% and a ROC AUC of around 90\%. These findings suggest that our approach effectively captures the underlying utility function from pairwise comparisons, supporting its potential for practical applications in drug candidate screening.

\begin{table}[t!]
\centering
\begin{tblr}{
  column{2} = {c},
  column{3} = {c},
  hline{1-2,8-9} = {-}{},
}
           & \textbf{Accuracy (\%)} & \textbf{ROC AUC}     \\
ackley     & 95.75 $\pm$ 1.97       & 0.99 $\pm$ 0.01      \\
alpine1    & 79.73 $\pm$ 2.99       & 0.88 $\pm$ 0.03      \\
hartmann   & 90.52 $\pm$ 2.18       & 0.98 $\pm$ 0.01      \\
levy       & 94.22 $\pm$ 1.31       & 0.99 $\pm$ 0.01      \\
dropwave   & 66.32 $\pm$ 3.89       & 0.72 $\pm$ 0.05      \\
qeifail    & 95.95 $\pm$ 1.35       & 0.99 $\pm$ 0.01      \\
human      & 80.40 $\pm$ 0.03       & 0.90 $\pm$ 0.02      \\
\end{tblr}
\caption{GP Utility Performance across 20 trials on synthetic and human data using 80/20 split of 1200 pair comparisons, demonstrating that the model effectively captures the latent utility.}
\label{tab:preference_results}
\end{table}

\begin{tcolorbox}[colback=green!10,boxsep=1mm,arc=1mm,outer arc=1mm]
\textbf{Summary}: Preferential learning robustly recovers the latent utility function with high accuracy and AUC on both synthetic and human data.
\end{tcolorbox}

\subsection{\mbox{Accuracy-Efficiency Trade-off in Measurement}}
\label{sec:affinity_measure}
\begin{figure}[t!]
    \centering
    \includegraphics[width=0.49\textwidth]{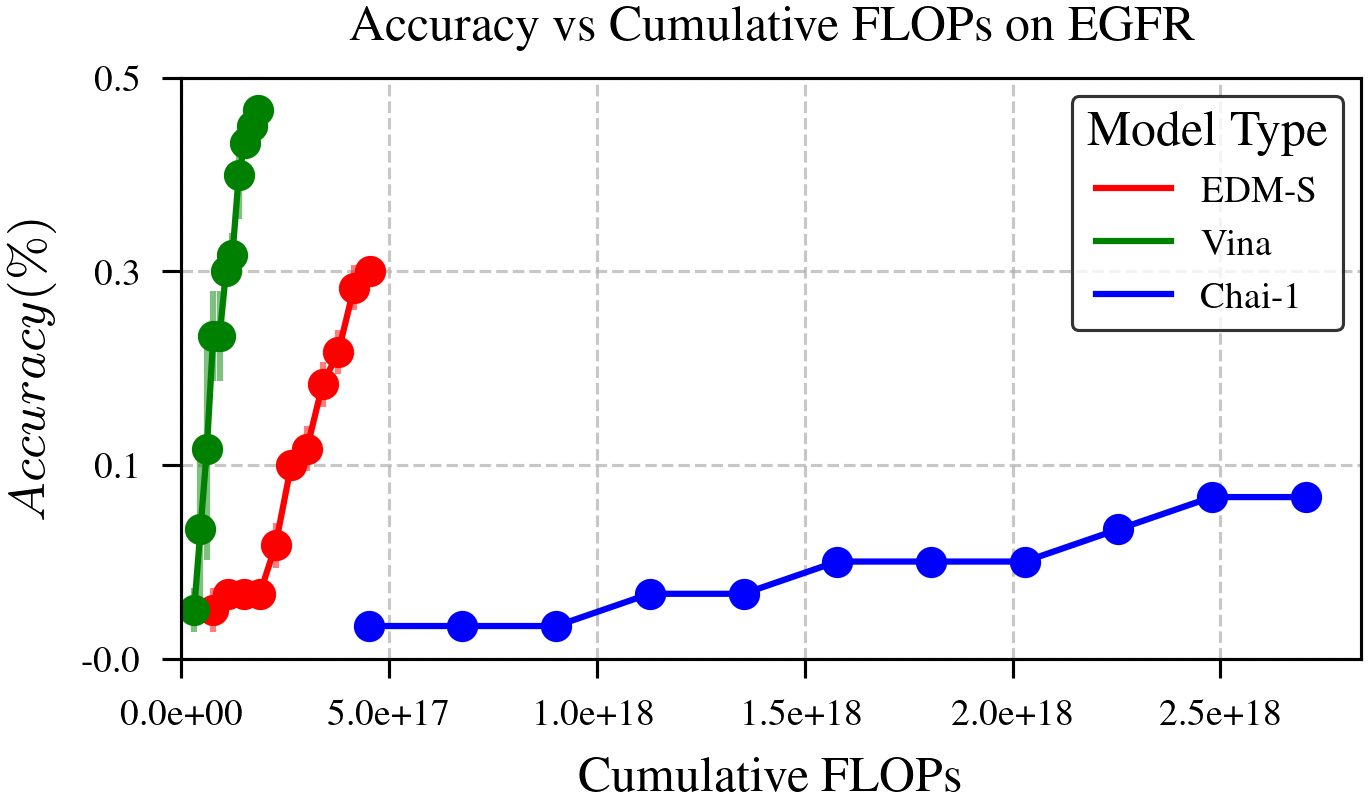}
    \caption{Accuracy over cumulative FLOPs on EGFR under the same screening settings. Vina achieves the highest accuracy with the fewest FLOPs, EDM-S is in between, and Chai uses the most FLOPs with the lowest accuracy.}
    \label{fig:acc_flops}
\end{figure}

We tackle \textit{RQ2} by investigating how the computational overhead of diffusion models affects convergence speed. We focus on the EGFR target, plotting accuracy against total FLOPs (FLoating OPerations). We first compare Chai-1 and Vina under the $\epsilon$-greedy acquisition function, noting that Chai produces five poses and Vina produces ten. Figure \ref{fig:acc_flops} shows that Vina requires fewer FLOPs to reach the highest accuracy of around 0.43, while Chai-1 is FLOP-intensive and ends with the lowest accuracy, around 0.10. These results show that slower, computationally expensive docking methods impede the exploration-exploitation cycle, while faster models (Vina) significantly boost throughput and convergence speed. Minimizing docking overhead enables the exploration of a larger chemical space.

Understanding the importance of efficiency in VS, we aim to train a highly efficient diffusion model while maintaining its accuracy using data augmentation. Our model (EDM-S) is based on Karras's diffusion model \cite{edm}. Training occurs in two phases: We first pre-train on 11 million synthetic pairs generated via pharmacophore alignment to capture diverse docking patterns. This enables the model to learn generalizable features from a broad chemical space that would be difficult to obtain solely from experimental data. We then fine-tune on PDBScan22, a high-quality dataset of approximately 180,000 experimentally determined complexes, to refine the model’s understanding of biologically relevant interactions. For targeted applications in (\ref{sec:human_study}), EDM-S is further fine-tuned on 10,000 pairs from \citet{garcia2022dockstring}, ensuring robust binding affinity predictions. EDM-S achieves a final accuracy of about 0.30, drastically improving over Chai-1 but still lagging behind Vina. This result shows that diffusion models can be efficient via data augmentation, and future research should investigate methods to make these models even more practical for VS. 

We further discuss the diffusion training result to deepen the understanding of the link between the training process and the downstream VS performance. Binding affinity (measured in kilocalories per mole) is evaluated using EGFR as the target, and results from all VS experiments are collected using the Vinardo scoring function. As shown in Figure \ref{fig:box_affinity}, EDM-S outperforms Chai in binding affinity measurements, highlighting its advantage in robust sampling. While EDM-S is slightly outperformed by Vina, it achieves comparable binding affinity distributions, showcasing its ability to balance accuracy and speed of measuring binding affinity. However, these findings are specific to the EGFR target; further experiments on diverse protein systems are necessary to assess broader generalizability.

\begin{figure}[t!]
    \centering
    \includegraphics[width=0.49\textwidth]{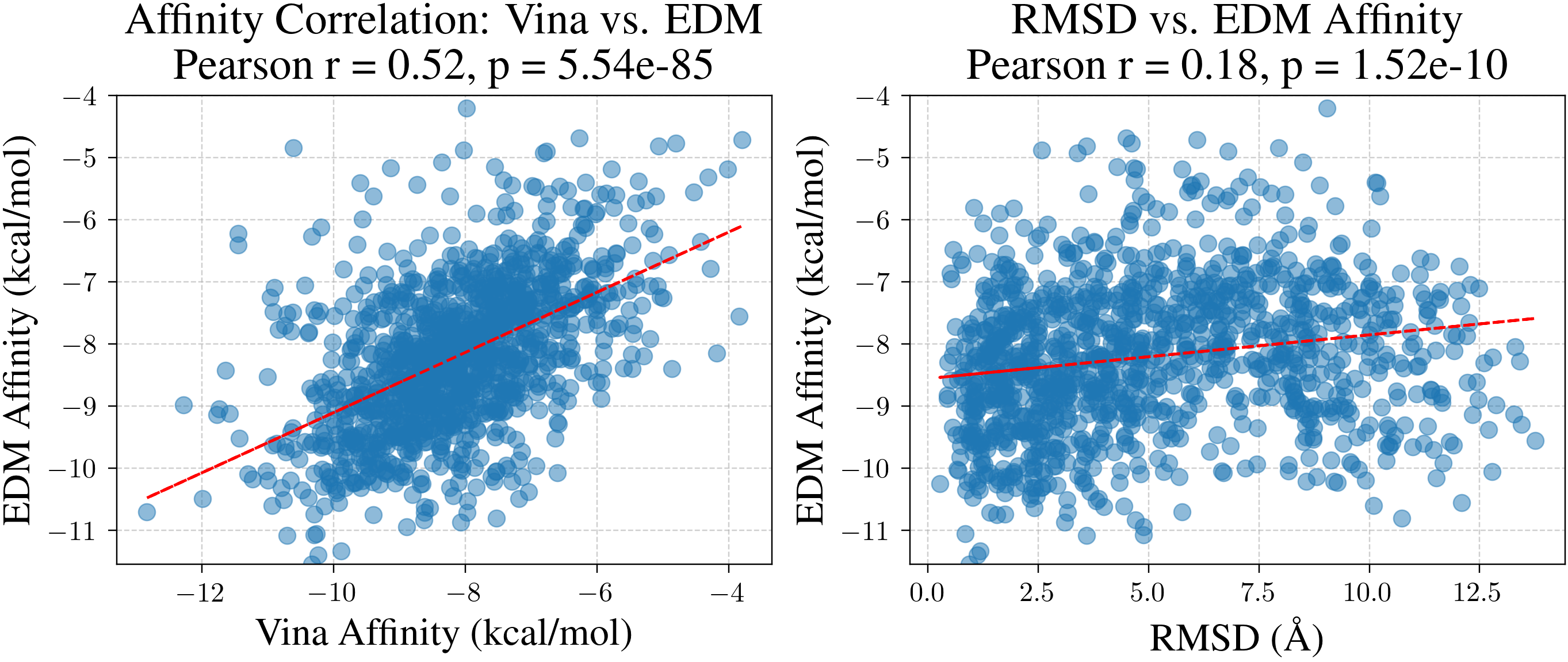}
    \caption{Scatter plots comparing EDM Affinity with Vina Affinity (left) and RMSD (right). A moderate correlation is observed between EDM and Vina affinities (r = 0.52), while no meaningful correlation exists between RMSD and EDM Affinity (r = 0.18).}
    \label{fig:affinity_rmsd}
\end{figure}

Regarding Root Mean Square Deviation (RMSD) performance, EDM-S, and DockScan22 (a DiffDock-S trained on our PDBScan22) employ distinct training strategies. RMSD, measured in {\AA}ngstr{\"o}ms (\AA, where 1\AA{} = 0.1 nm), quantifies structural deviations between predicted and reference ligand poses, with lower values indicating higher accuracy. EDM-S combines pretraining on the PapyrusScan dataset (11M synthetic pairs) with fine-tuning on PDBScan22 (322K validated complexes), while DockScan22 trains solely on PDBScan22 using DiffDock-S as its backbone. Experimental results (\ref{tab:training_result}) show DockScan22 achieves 54.1\% and 34.1\% accuracy (RMSD $<$ 2\AA) on PoseBuster V1 and PDBBind, outperforming DiffDock-S and the original DiffDock. EDM-S achieves 91\% accuracy (RMSD $<$ 5\AA) on PoseBuster V1. Both models are 34 times faster than folding models like AlphaFold, running in 10s on an A100 GPU, demonstrating their practicality for large-scale applications. Most current methods train diffusion models on atomic coordinates, optimizing for low RMSD. However, RMSD only quantifies geometric similarity to a reference structure and is not sensitive to steric clashes or energetically unfavorable interactions: a ligand with RMSD $<$ 2\AA{} may still exhibit steric clashes that disrupt binding, making it a poor candidate despite its structural similarity. In drug discovery, binding strength is more important than geometric accuracy. While binding affinity reflects a ligand's ability to bind to a protein and regulate its function, geometric accuracy only indicates how closely the predicted pose aligns with a reference structure, offering little insight into the drug's regulatory potential. The Vinardo scoring function provides a more meaningful measure of binding affinity by incorporating both energetic and steric factors. Figure \ref{fig:affinity_rmsd} shows the weak RMSD-affinity correlation, emphasizing the need for affinity-based scoring as a better measure of drug quality.

\begin{tcolorbox}[colback=green!10,boxsep=1mm,arc=1mm,outer arc=1mm]
\textbf{Summary}: Diffusion models show promise in binding affinity prediction, though physics-based methods demonstrate greater efficiency and accuracy.
\end{tcolorbox}

\subsection{\mbox{Eliciting Expert Preference for Efficient Screening}}
\label{sec:human_study}

To address \textit{RQ3}, we focus on two targets, Epidermal Growth Factor Receptor (EGFR) and Dopamine D2 Receptor (DRD2) proteins, due to their clinical importance and the availability of multiple FDA-approved drugs~\cite{cohen2021kinasetimeline}. We collect 37 and 58 FDA-approved or late-stage clinical candidates from the PKIDB and DrugBank~\cite{carles2018pkidb, drugbank} for EGFR and DRD2, respectively, treating them as ``goal-optimal'' molecules. The screening library comprises $260,000$ molecules from \citet{garcia2022dockstring}, in which a random subset of $100,000$ is used to simulate a realistic campaign. Expert chemists provide preference labels, defining nuanced \emph{multi-objective} utility functions. In each BO iteration, these experts complete 200 pairwise comparisons via an interactive app, yielding a total of $2,200$ pairs over one full experiment of the study. Overall, the ranking process took a chemist roughly 10 hours to complete. During the process, they have access to SMILES visualizations and ligand properties to inform their decisions (see Figure \ref{fig:app_vs}). Our current experiment involves only one chemist; we leave bias mitigation for future work by incorporating feedback from multiple chemists. Finally, We also evaluate an affinity-optimization baseline to determine whether multi-property feedback yields more meaningful optimization for VS, using only Vina, as it typically provides the best performance for affinity-based docking.

As DRD2 is a protein located inside the Central Nervous System (CNS), its drug candidates require substantial considerations for brain penetration. To reflect the distinct pharmacological considerations, we select new objectives for DRD2 that differ from the previous target EGFR. For EGFR, we optimize affinity, molecular weight (MW), lipophilicity, and half-life, aligning with key properties of kinase inhibitors. For EGFR, these properties are vital as they enable potent target binding, efficient cell penetration, and sustained drug activity-key traits for effective kinase inhibition. Meanwhile, for DRD2, we instead optimize affinity, MW, topological polar surface area (TPSA), predicted drug-induced liver injury (DILI), and predicted blood-brain barrier permeability (BBB). MW, TPSA, and BBB reflect important parameters allowing brain permeability, while DILI provides a standard toxicity indicator; for details on target-specific objective selection, see Appendix~\ref{sec:guideline}.

To validate our objective selection, we compare these properties between drugs and non-drug molecules, confirming that the drug-like molecules exhibit characteristics consistent with our assumptions (see ~\cref{fig:drd2_lig,fig:egfr_lig}). These choices ensure that our preference optimization aligns with real-world drug design. The BO pipeline begins by randomly sampling $1.0\%$ of the $100,000$-compound library, then screening an additional $0.5\%$ per iteration for 10 iterations (covering $6\%$ of the library). All experiments are run on an A100 GPU with two seeds for EGFR and one for DRD2. Chai-1 requires 180 GPU-hours to complete 6000 dockings, making it computationally expensive for high-throughput tasks. In contrast, EDM-S finished in 17 GPU-hours. The BO computation--integrated within the cheapvs process--takes around 12 GPU-hours, resulting in an overall process time of about 3 days. Meanwhile, Vina requires no docking as affinities are precomputed from \citet{garcia2022dockstring}. However, for a fair comparison, we refer to the measurements from \citet{ding2023vina} for Vina on GPU, which we estimate a runtime of approximately 2.4 GPU-hours for 6,000 docking runs.

\begin{table}[t!]
\centering
\label{tab:model_performance}
\begin{tabular}{lcc}
\hline
\textbf{Model} & \textbf{ROC AUC} & \textbf{Accuracy} \\
\hline
No Interactions & 0.67 $\pm$ 0.21 & 0.61 $\pm$ 0.16 \\
Second-Order    & 0.72 $\pm$ 0.14 & 0.68 $\pm$ 0.11 \\
Third-Order       & 0.77 $\pm$ 0.11 & 0.71 $\pm$ 0.09 \\
Fourth-Order   & \textbf{0.79 $\pm$ 0.08} & \textbf{0.74 $\pm$ 0.05} \\
\hline
\end{tabular}
\caption{Linear regression performance across interaction orders. No Interaction uses individual features, while Pairwise, Triple, and Quadruple add second-, third-, and fourth-order interactions. Results averaged over 20 trials with 1200 pair-wise expert preferences (80/20 split), suggest trade-offs in ligand properties.}
\label{tab:interation_table}
\end{table}

Figure~\ref{fig:moavs_hp} shows how effectively each approach (Multi-Objective, Affinity-Only, and Random) with different dock models (Vina, Chai, EDM-S) identifies the known EGFR and DRD2 ligands. For EGFR experiments, using Vina strategy, guided by expert preferences, attains about 42\% accuracy in retrieving these known drugs, substantially surpassing the 22\% accuracy of the best affinity-only approach. EDM-S reaches up to 30\%. Chai-1 performs poorly due to its high affinity, highlighting that affinity still remains a crucial component in multi-objective optimization. Random screening performs poorly. We observe that two docking models show improved performance when incorporating multi-objective preferences over single-objective affinity, emphasizing the broad advantage of reflecting real-world trade-offs in the BO process. The results for DRD2 show that our multi-objective approach identifies a greater fraction of the 58 known DRD2 drugs compared to the affinity-only model and random selection. After screening about 1200 ligands, its accuracy quickly rises above 60\%, while the best affinity-only model remains at zero. These findings \textit{address RQ3}: leveraging expert preference leads to more clinically relevant molecules than relying solely on affinity, and reinforces our hypothesis that incorporating expert-defined preferences leads to more effective VS

\begin{figure}[t!]
  \centering
  \includegraphics[width=0.49\textwidth]{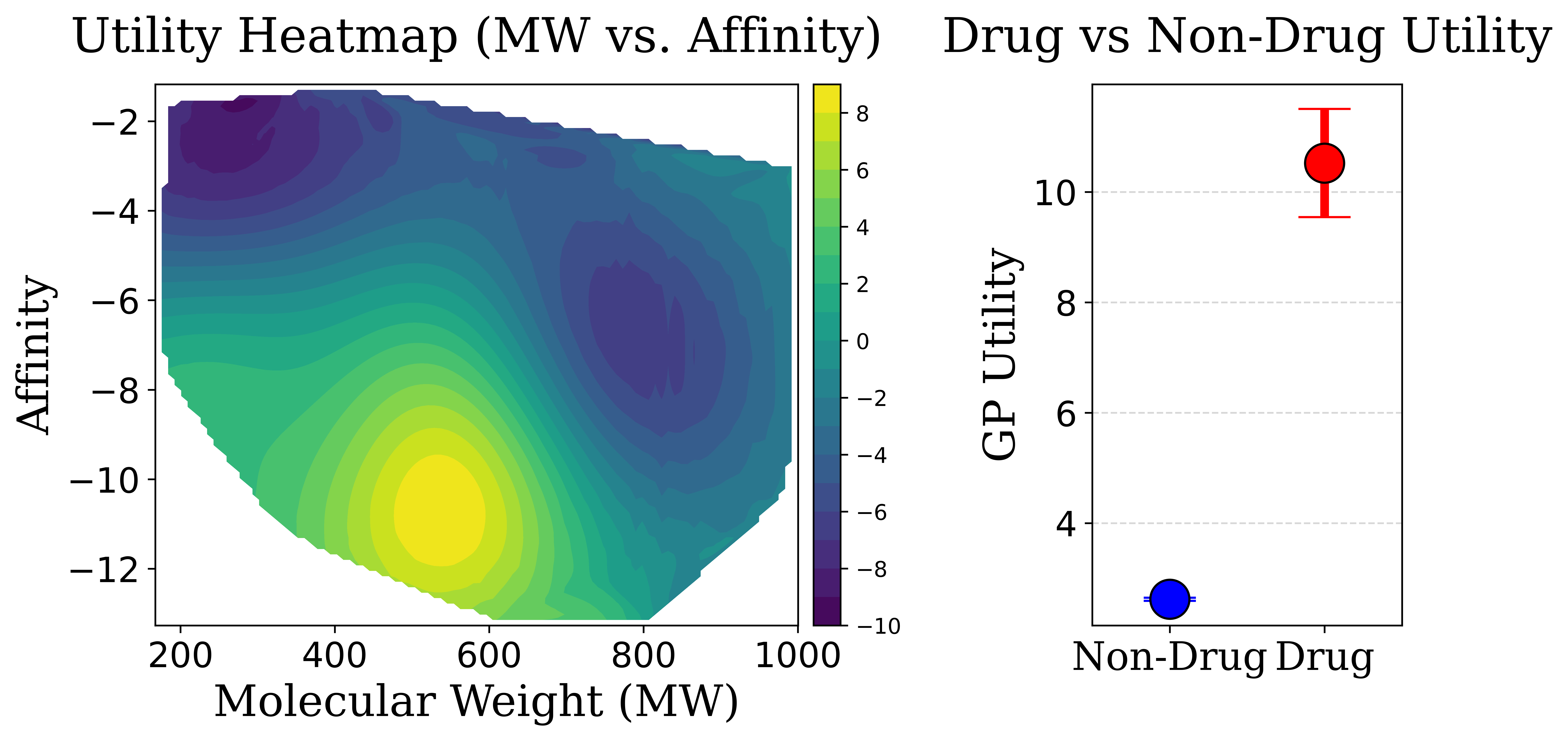}
  \caption{Predictive utility scores after BO on expert preference elicitation. Heatmaps show utility for two objectives (others fixed at the mean), while box plots compare the mean scores of drugs vs. non-drugs with 95\% CI bars, highlighting the algorithm captures domain knowledge and balances competing properties.}
  \label{fig:elicitation}
\end{figure}

To understand why multi-objective optimization is more effective, we analyze how the utility model captures expert preferences and the resulting trade-offs in ligand selection. Figure~\ref{fig:elicitation} illustrates that the utility model captures expert preferences on EGFR: the box plot shows higher mean utility scores for drug-like compounds, while the heatmap highlights the \textit{trade-offs} of multi-objective optimization. This interplay between pharmacokinetic properties reinforces how the model balances trade-offs to identify clinically relevant candidates. Understanding the trade-off nature of drug discovery requires understanding how optimizing one objective impacts others. Toward this goal, we use simple linear regression to test whether high-order interaction is necessary for out-of-sample fit. The null model consists of individual effects, while the alternative models incorporate higher-order interactions to capture the interdependence among ligand properties. Table~\ref{tab:interation_table} shows that models with higher-order interactions generalize better, indicating that ligand properties exhibit complex interdependencies that influence predictive performance. GP, our main predictive model, generalizes this ideal to infinite dimensional feature space, capturing high-order interaction terms through kernel functions \cite{scholkopf2002learning, mercer1909functions, williams1998computation}, allowing it to naturally model intricate dependencies among ligand properties without explicitly defining interaction orders. The superior out-of-sample fit of the alternative models and the complex utility landscape conclude that optimizing one objective does not necessarily improve overall drug potential, highlighting the shortcomings of single-objective screening.

\begin{tcolorbox}[colback=green!10,boxsep=1mm,arc=1mm,outer arc=1mm]
\textbf{Summary}: Incorporating expert preferences outperforms affinity-only methods, emphasizing the critical role of chemical intuition in drug discovery.
\end{tcolorbox}

\section{Conclusion}
We present a framework for accelerating drug discovery with preferential multi-objective BO. \framework{} enables a deeper understanding of how incorporating chemical intuition can enhance the practicality of the VS. By addressing the challenges chemists often face during hit identification, \framework{} speeds up the VS process. It requires screening only a small subset of the ligand library and leveraging a few chemists' pairwise preferences to efficiently identify drug-like compounds. Specifically, \framework{} successfully identified up to 16 out of 37 known drugs for EGFR and 36 out of 57 for DRD2 targets. This paper opens exciting avenues for future research. \framework{} relies on pairwise preference and is well-suited for listwise preference. Here, chemists can select the best ligand from a list, providing richer preference information and further boosting algorithm performance. Future work would benefit from exploring advanced preference modeling to enable deeper insights and further accelerate the drug discovery process.

\section{Impact Statement}
This work advances Preferential Multi-Objective BO in drug discovery, enhancing the efficiency of identifying promising therapeutic compounds. The potential societal benefits include accelerating the identification of high-priority drug candidates, which may contribute to advancements in healthcare and therapeutic development. Ethical considerations were taken into account, particularly in designing experiments that reflect real-world decision-making while minimizing computational and resource biases. We do not foresee any immediate negative societal consequences, but we encourage further discussion as the field progresses and practical applications emerge.

\section{Acknowledgement}
This work was supported by the 2024 HAI-Google Cloud Credits Grant on ``Proactive Pandemic Preparedness: Accelerating Antiviral Drug Discovery by Empowering Chemists with Deep Generative Models.'' LH.P. acknowledges support from the Vingroup Science and Technology Scholarship for Doctoral Degrees and EPSRC 2886971. SK acknowledges support by NSF 2046795 and 2205329, IES R305C240046, ARPA-H, Stanford HAI, RAISE Health, OpenAI, and Google.

\bibliographystyle{plainnat}
\bibliography{references}
\clearpage
\appendix
\onecolumn
\section{Notation}
We summarize the notation used in our paper in Table~\ref{notation}.

\begin{table}[hbt!]
\centering
\begin{tblr}{
  hline{1-2} = {-}{},
  hline{18} = {-}{0.08em},
  colspec = {c p{10cm}}, 
}
\textbf{Symbol} & \textbf{Description}                                                                    \\
$\Lc$           & Ligand library used for VS.                                              
                    \\
$\ell_i$           & Ligand $i$ in the ligand library $\Lc$.                                                                          \\
$\ell_{\text{3D}}$      & 3D coordinate vector of ligand $\ell$.                                      
                        \\

$g_P$             & Affinity model mapping ligand fingerprints to binding affinity.
                        \\
$f$             & Latent Utility model learning from preference data.               
                        \\
$h$             & Physics-based affinity scoring function.               
                        \\
$x$             & Ligand properties, including physicochemical and ADMET.
                        \\
$\ell_\Mc$             & Morgan Fingerprint representation of the ligand's structure.                                                      \\
$\alpha$        & Acquisition function in BO for ligand selection.                \\
$R$             & Regret, quantifying the gap between the best possible and selected ligand.        
                        \\
$U$             & Utility values of ligands.                                      \\
$k$             & Used for selecting the top-$k$ compounds.                                               \\
$\rho$             & Protein target for VS.                                                   \\
$p_\theta$        & Docking diffusion model, predicting ligand-protein binding. 
                \\
$\Dc_{g_{\rho}}$    & Datasets acquired to train affinity model. 
                \\
$\Dc_{f}$           & Datasets acquired to train utility model. 

\end{tblr}
\caption{Notation}
\label{notation}
\end{table}

\section{Acquisition Functions}
\label{sec:appendix_acq}
In this paper, we utilize the following acquisition functions to guide our optimization process:

\begin{itemize}
    \item \textbf{qExpected Improvement (qEI)} \cite{EI}: Evaluates the expected gain in model performance across multiple candidates, emphasizing exploration where improvement potential is high.
    \item \textbf{qProbability of Improvement (qPI)} \cite{PI}: Computes the likelihood that a set of candidate samples will surpass the current best performance.
    \item \textbf{qUpper Confidence Bound (qUCB)} \cite{UCB}: Balances exploration and exploitation by selecting candidates with both high uncertainty and high predicted performance based on their upper confidence bounds.
    \item \textbf{qThompson Sampling (qTS)} \cite{TS}: Approximates the posterior distribution of the model and sample candidates to maximize predicted utility, promoting diverse exploration.
    \item \textbf{qExpected Utility of the Best Option (qEUBO)} \cite{qeubo}: A decision-theoretic acquisition function for preferential BO (PBO) that maximizes the expected utility of the best option. It is computationally efficient, robust under noise, and offers superior performance with guaranteed regret convergence.
    \item \textbf{Greedy}: Selects the candidate with the highest predicted performance at each step. This purely exploits the current model estimates and does not explicitly encourage exploration.
    \item \textbf{\(\epsilon\)-Greedy} \cite{lai1985asymptotically}: With probability \(\epsilon\) (typically 5\%), selects a candidate at random (exploration), and otherwise selects the best predicted candidate (exploitation). This method is a simple yet effective way to balance exploration and exploitation.

\end{itemize}

\section{Preliminary Analysis on the Data}
\label{sec:appendix_preliminary_analysis}
\begin{figure}[hbt!]
\centering
\includegraphics[width=0.9\textwidth]{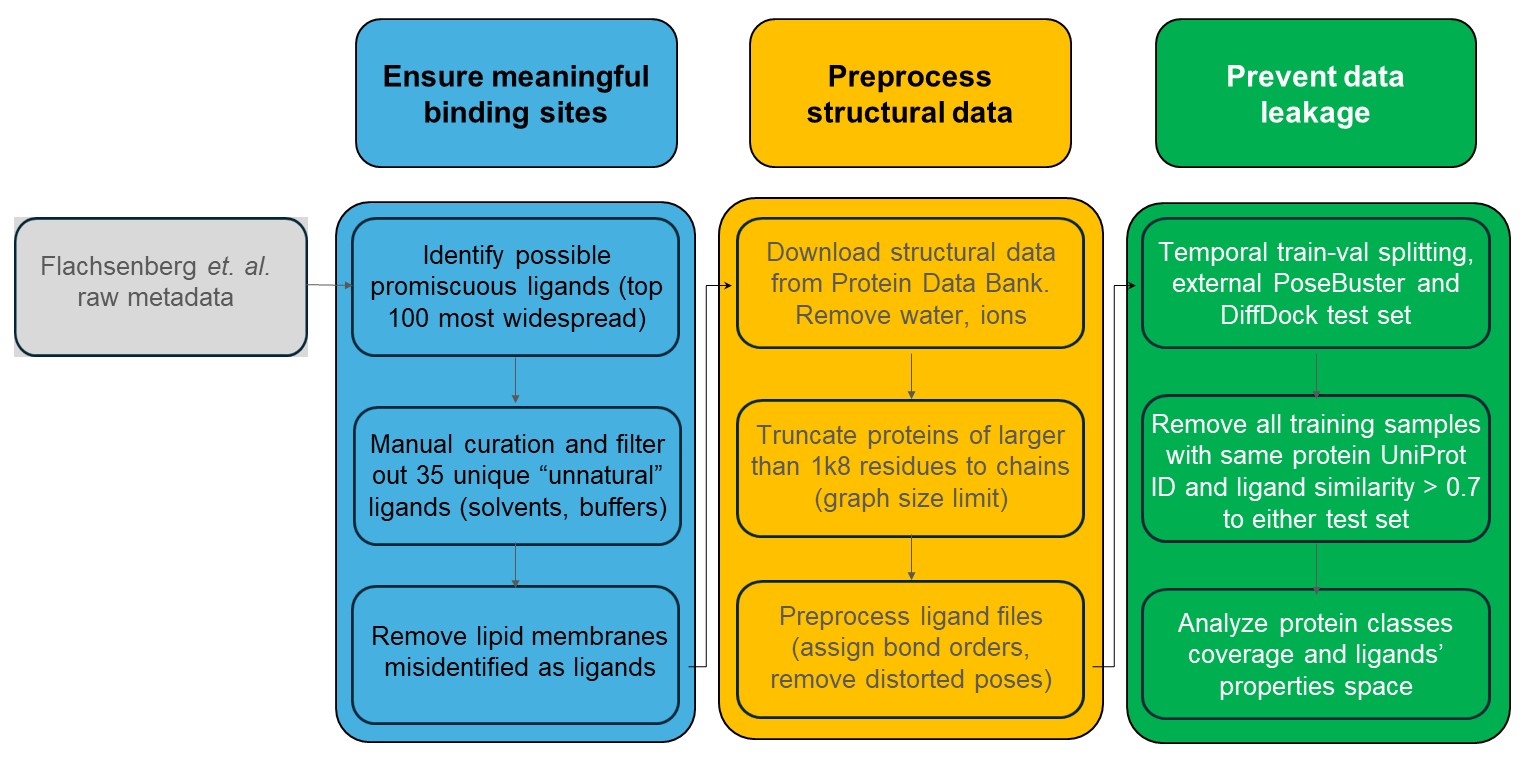}
\caption{PDBScan22 data curation workflow. The process consists of three main steps: (1) Ensuring meaningful binding sites by filtering promiscuous ligands, removing unnatural molecules such as solvents and buffers, and eliminating misidentified lipid membranes. (2) Preprocessing structural data by downloading structures from PDB, removing water and ions, truncating proteins exceeding 1,800 residues, and refining ligand files by assigning bond orders and eliminating distorted poses. (3) Preventing data leakage through temporal train-validation splitting, and removing training samples with proteins sharing UniProt IDs and highly similar ligands (\textgreater0.7 similarity) with test sets.}
\label{fig:image4}
\end{figure}

As noted in Figure \ref{fig:proteins}, the number of data points in the PDBScan training data is roughly four times as large as the data points in the PDBbind training data. Furthermore, the training data utilized covers 18 different protein groups. Additionally, we also perform a similar comparison on the Plinder dataset \cite{plinder} to further evaluate the differences in data distribution and model performance across diverse datasets.

\subsection{Diversity of Data: Proteins}
\begin{figure}[!hbt]
    \centering
    \includegraphics[width=0.49\textwidth]{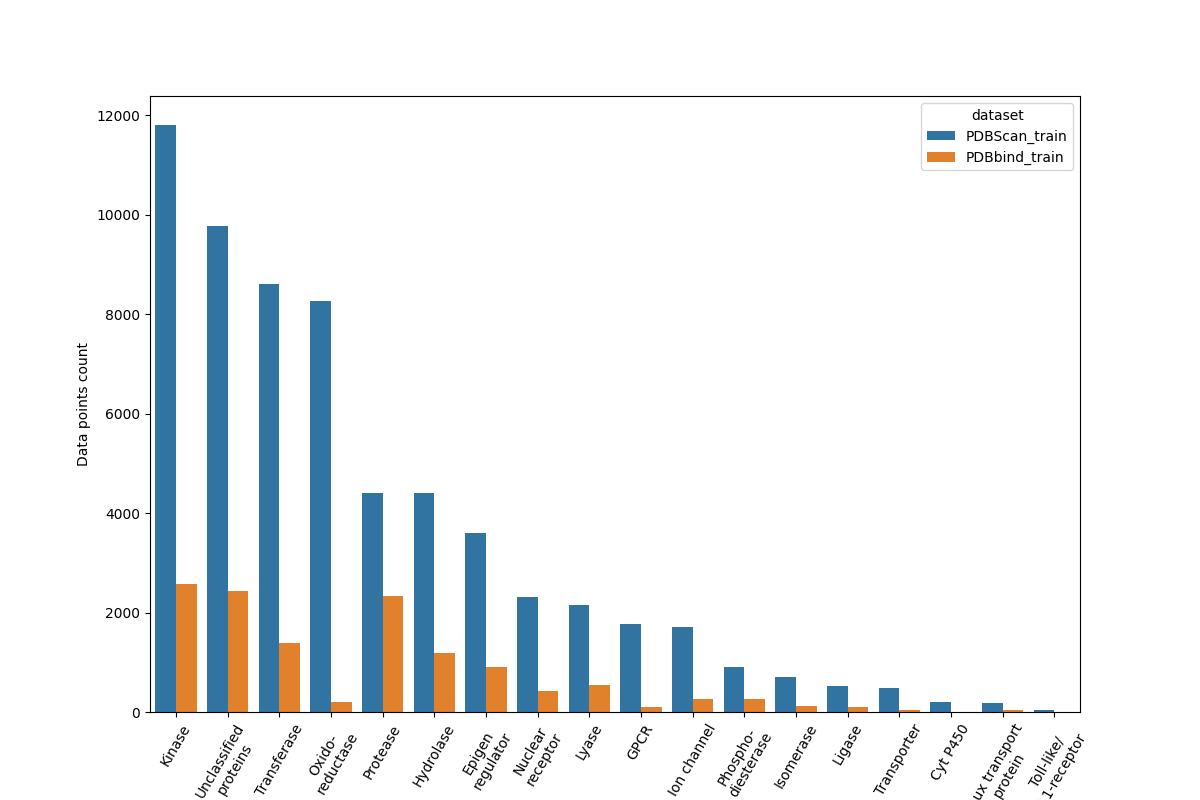}
    \includegraphics[width=0.45\textwidth]{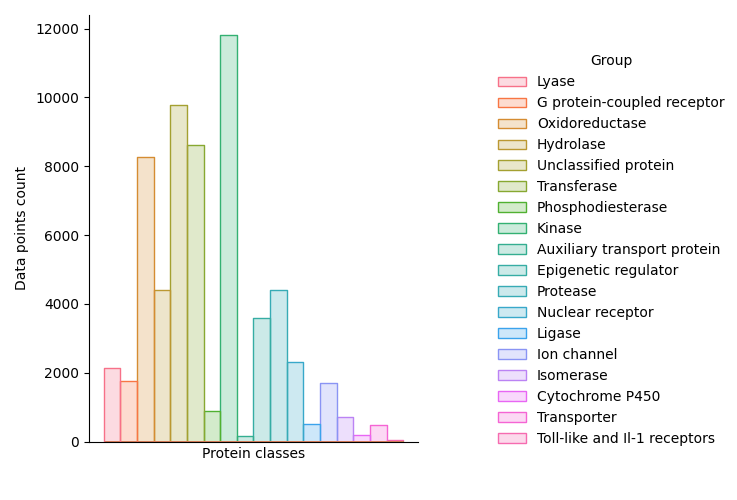}
    \caption{Analysis of protein properties: Protein classes distribution, Protein classes data point count}
    \label{fig:proteins}
\end{figure}
Understanding the diversity of protein classes in the dataset is essential for evaluating its coverage and potential biases in molecular docking tasks. Figure~\ref{fig:proteins} illustrates the distribution of protein classes across different datasets, highlighting variations in data availability. The left panel compares the protein class distributions between PDBScan and PDBbind, showing that PDBScan contains a significantly larger number of data points across all protein categories, particularly in ``Unclassified proteins'' and ``Kinases.'' This discrepancy suggests that PDBScan provides broader protein coverage, which may enhance model generalization.

The right panel further details the absolute counts of protein classes, emphasizing their relative abundance. The dataset is dominated by enzymatic proteins, including Oxidoreductases, Transferases, and Hydrolases, which are frequently studied in drug discovery. However, certain categories such as Toll-like and IL-1 receptors, Transporters, and Cytochrome P450 remain underrepresented, potentially impacting model performance on these classes. These insights highlight the importance of data augmentation techniques to balance protein representation and improve downstream learning.

\subsection{Diversity of Data: Ligands}
\begin{figure}[H]
    \centering
    \includegraphics[width=0.9\textwidth]{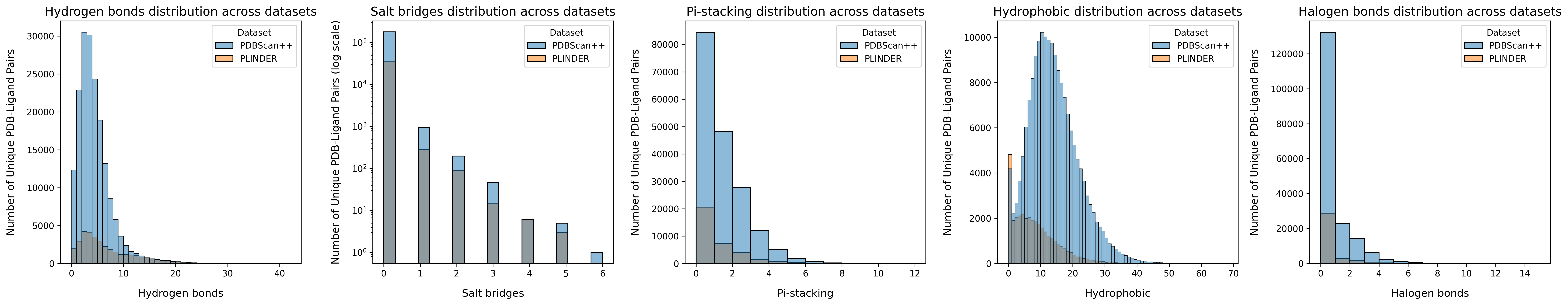}
    \caption{Distribution of molecular interactions in PDBScan++ and PLINDER, including hydrogen bonds, salt bridges, pi-stacking, hydrophobic interactions, and halogen bonds. PDBScan++ exhibits a broader range of interactions due to its larger dataset size.}
    \label{fig:interation}
\end{figure}

Figure \ref{fig:interation} compares the distribution of key molecular interactions across the PDBScan++ and PLINDER datasets, including hydrogen bonds, salt bridges, pi-stacking, hydrophobic interactions, and halogen bonds. PDBScan++ consistently contains more ligand-protein interactions than PLINDER, reflecting its larger dataset size. Hydrogen bonds and hydrophobic interactions are the most prevalent, while halogen bonds are the least common. Notably, PDBScan++ includes PDBScan22 along with 250k synthetic pharmacophore-ligand pairs with the lowest affinity, added to match the number of compounds in PLINDER, ensuring a balanced comparison.

\begin{figure}[H]
    \centering
    \includegraphics[width=1.0\textwidth]{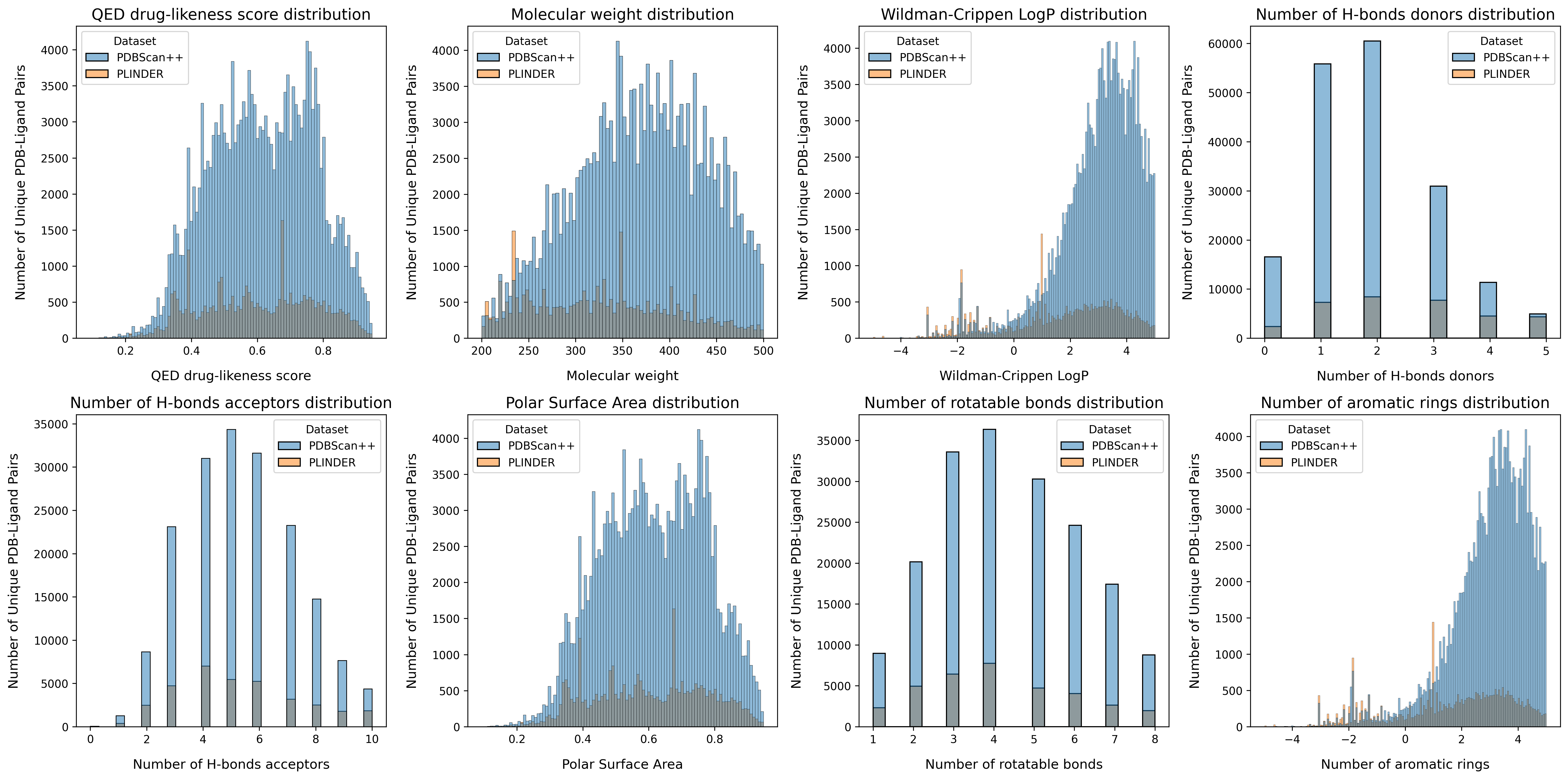}
    \caption{Comparison of physicochemical properties between PDBScan++ and PLINDER, including QED drug-likeness, molecular weight, LogP, hydrogen bond donors/acceptors, polar surface area, rotatable bonds, and aromatic rings. PDBScan++ shows greater diversity across all properties.}
    \label{fig:physchem}
\end{figure}

Figure \ref{fig:physchem} presents the distribution of key physicochemical properties across the PDBScan++ and PLINDER datasets, including QED drug-likeness, molecular weight, Wildman-Crippen LogP, hydrogen bond donors and acceptors, polar surface area, rotatable bonds, and aromatic rings. Across all properties, PDBScan++ exhibits a broader and more diverse range of molecular characteristics compared to PLINDER, reflecting its larger dataset size. The QED scores and molecular weights of compounds in both datasets follow similar distributions, but PDBScan++ has a wider spread. The Wildman-Crippen LogP distribution indicates that PDBScan++ includes more hydrophobic molecules. Additionally, PDBScan++ contains a higher number of hydrogen bond donors and acceptors, as well as greater structural flexibility (rotatable bonds) and aromaticity (aromatic rings), highlighting its increased chemical diversity.

Together with the molecular interaction distributions in Figure \ref{fig:interation}, these results emphasize that PDBScan++ encompasses a broader and more chemically diverse set of compounds than PLINDER, ensuring a comprehensive representation of molecular properties. Notably, PDBScan++ includes PDBScan22 along with 250k synthetic pharmacophore-ligand pairs with the lowest affinity, introduced to match the number of compounds in PLINDER.

\subsection{Diversity of Protein-Ligand Dataset}
\begin{table}[H]
    \centering
    \setlength\extrarowheight{3pt}
    \begin{tabular}{p{3cm} p{2cm} p{4.5cm} p{4.5cm}}
    \hline
    \textbf{Dataset} & \textbf{Dataset size} & \textbf{Main approaches} & \textbf{Limitations} \\ \hline
       PDBBind \citep{liu:17}  & Approx. 20k complexes & Human-curated experimental structures with experimental binding affinity & Limited number of data points \\ 
       BindingMOAD \citep{wagle2023bindingMOAD} & Approx. 40k complexes & Human-curated experimental structures (with or without binding affinity) & Limited number of data points \\
       BioLiP2 \citep{zhang2024biolip2} & Approx. 470k organic ligand complexes & Semi-manual-curated experimental structures (with or without binding affinity) & Include non-specific ligands (anions, crystal artefacts, solvents, etc) \\
       PDBScreen \citep{cao2024pdbScreen} & True ligands: 23k unique ligands \newline Generated decoys: approx. 110k & - Automated filtering from the PDB, excluding endogenous ligand (ATP, ADP, etc.) \newline - Data augmentation by redocking and cross-docking \newline - Also include artificially generated decoy ligands & - Purposely built for screening and scoring, but excluded endogenous ligands which represent important pockets \newline - Redocked and cross-docked poses do not enrich protein or ligand diversity\\
       PLINDER \cite{plinder} & Approx. 450k unique (organic) ligands & - Automated annotation of PDB structures to retrieve broad-termed ligands \newline - Graph-based multiple-similarity data splitting to debias the train-test split & - Include covalent modifications of the proteins as surrogate ligands (glycosylation) \newline - Include ions on the broadly defined ligand category\\
       PapyrusScan  & Approx. 11 million & - Synthetic data from 2D binding information & - Synthetic data \newline - Unbalanced number of data points between proteins depending on 2D data\\ \hline
    \end{tabular}
    \caption{Comparison of community-available protein-ligand structural datasets.}
    \label{tab:table2}
\end{table}

Table \ref{tab:table2} compares various protein-ligand structural datasets, with a focus on PDBScreen and PapyrusScan in contrast to existing community datasets. PDBScreen refines structural data by filtering endogenous ligands and augmenting diversity through redocking and cross-docking, making it well-suited for screening and scoring tasks. However, its exclusion of endogenous ligands may overlook important binding pockets. In contrast, PapyrusScan is the largest dataset, containing approximately 11 million protein-ligand interactions derived from 2D binding data, offering extensive coverage but relying on synthetic data, leading to potential biases and imbalances across proteins. Compared to PDBBind, BindingMOAD, and BioLiP2, which primarily rely on human-curated or semi-curated experimental structures, PDBScreen and PapyrusScan emphasize data augmentation and large-scale synthetic generation, respectively. While PLINDER provides a broad dataset with automated annotation and debiased train-test splitting, it includes covalent modifications and ions as ligands, introducing potential noise. This comparison highlights the complementary nature of PDBScreen and PapyrusScan, balancing curated experimental data with large-scale synthetic augmentation to enhance ligand-protein modeling.

\section{More Results on Preferential Multi-Objective Bayesian Optimization}
\label{sec:MOAVS_accuracy}

\subsection{CheapVS's Pseudocode}
\label{sec:pseudocode}
\begin{algorithm}
\caption{CheapVS's Algorithm}
\begin{algorithmic}[1]
\REQUIRE Ligand library $\mathcal{L}=\{\ell_1,\dots,\ell_N\}$, target protein $\rho$, docking model $p_\theta$, acquisition function $\alpha$
\ENSURE Top-$k$ drug ligands for target $\rho$

\STATE $\mathcal{D} \gets \emptyset$, $\mathcal{D}_{g_{\rho}} \gets \emptyset$, $\mathcal{D}_f \gets \emptyset$, $\mathcal{F} \gets \emptyset$, $\mathcal{L}_{\text{tox}} \gets \emptyset$, $\mathcal{L}_{\text{sol}} \gets \emptyset$, $X_{\ell,\rho}^{\text{aff}} \gets \emptyset$
\STATE $\mathcal{D}_i \gets \{\ell \in \mathcal{L} \mid U(0,1) < 0.01\}$ \hfill // Select 1\% of $\mathcal{L}$ at random

\FOR{each ligand $\ell_i \in \mathcal{L}$}
    \STATE $\ell_{i,\mathcal{M}} \gets \textsc{MorganFingerprint}(\ell_i)$
    \STATE $x_{\ell_i}^{\text{tox}} \gets \textsc{RDKitToxicity}(\ell_i)$
    \STATE $x_{\ell_i}^{\text{sol}} \gets \textsc{RDKitSolubility}(\ell_i)$
    \STATE $\mathcal{F} \gets \mathcal{F} \cup \{ \ell_{i,\mathcal{M}} \}$, $\mathcal{L}_{\text{toc}} \gets \mathcal{L}_{\text{toc}} \cup \{ x_{\ell_i}^{\text{tox}} \}$,  $\mathcal{L}_{\text{sol}} \gets \mathcal{L}_{\text{sol}} \cup \{ x_{\ell_i}^{\text{sol}} \}$

\ENDFOR

\WHILE{computational budget not reached}
    \STATE $g_P \sim \mathcal{GP}(\mu, k)$ \hfill // Initialize affinity model with Gaussian likelihood
    \STATE $f \sim \mathcal{GP}(\mu, k)$ \hfill // Initialize utility model with pairwise likelihood:
    \STATE $\mathcal{D} \gets \mathcal{D} \cup \mathcal{D}_i$ \hfill // Add selected ligands to the dataset
    \STATE $\mathcal{L} \gets \mathcal{L} \setminus \mathcal{D}_i$ \hfill // Remove selected ligands from the library    
    
    \STATE \textbf{For} each ligand $\ell_i \in \mathcal{D}_i$:
    \STATE \quad $\ell_{i;\text{3D}} \sim p_\theta(\ell_i, \rho)$
    \STATE \quad $x_{\ell_i,\rho}^{\text{aff}} \gets \min_{\ell_{i;\text{3D}} \in \mathbb{R}^{3 \times N_{\ell_i}}} h(\ell_{i;\text{3D}},\rho)$
    \STATE \quad $X_{\ell,\rho}^{\text{aff}} \gets X_{\ell,\rho}^{\text{aff}} \cup \{ x_{\ell_i,\rho}^{\text{aff}} \}$

    \STATE $\mathcal{D}_{g_{\rho}} \gets \mathcal{D}_{g_{\rho}} \cup \{ (\mathcal{F}(\mathcal{D}_i),\, X_{\ell,\rho}^{\text{aff}}) \}$
    \STATE Fit $g_P$ on $\mathcal{D}_{g_{\rho}}$

    \STATE $\mathcal{I} \gets \text{random pairs}(\mathcal{D}_i)$ \hfill // $\mathcal{I}$: set of index pairs from $\mathcal{D}_i$
    \STATE $X_{\text{train}} \gets \text{concat}\Bigl(X_{\ell,\rho}^{\text{aff}},\mathcal{L}_{\text{tox}}(\mathcal{D}_i),\, \mathcal{L}_{\text{sol}}(\mathcal{D}_i)\Bigr)$

    \STATE $Y_{e} \gets \text{chemists\_ranking}(X_{\text{train}}, \mathcal{I})$
    \STATE $\mathcal{D}_f \gets \mathcal{D}_f \cup \{ (X_{\text{train}},\, Y_{e}) \}$
    \STATE Fit $f$ on $\mathcal{D}_f$

    \STATE $\hat{X}^{\text{aff}}_{\ell, \rho} \gets g_\rho(\mathcal{F}(\mathcal{L}))$ \hfill // Posterior inference: $\hat{x}^{\text{aff}}_{\ell, \rho} = g_\rho(\ell_\mathcal{M})\ \forall\ \ell \in \mathcal{L}$
    \STATE $\hat{X} \gets \text{concat}(\hat{X}^{\text{aff}}_{\ell,\rho},\, \mathcal{L}_{\text{sol}}(\mathcal{L}),\, \mathcal{L}_{\text{tox}}(\mathcal{L}))$
    \STATE $\mathcal{D}_i \gets \text{Top}_k \Bigl\{ \ell \in \mathcal{L} \mid \Ebb_{p(x_{\ell, \rho}^{\text{aff}} \mid \ell, \Dc_{g_\rho})} \alpha(f(\hat{X})) \Bigr\}$

\ENDWHILE
\STATE \textbf{Return} $\mathcal{D}$
\end{algorithmic}
\end{algorithm}

\subsection{DRD2 Experiments}
\label{sec:drd2}
\begin{figure}[H]
\centering
\includegraphics[width=0.90\textwidth]{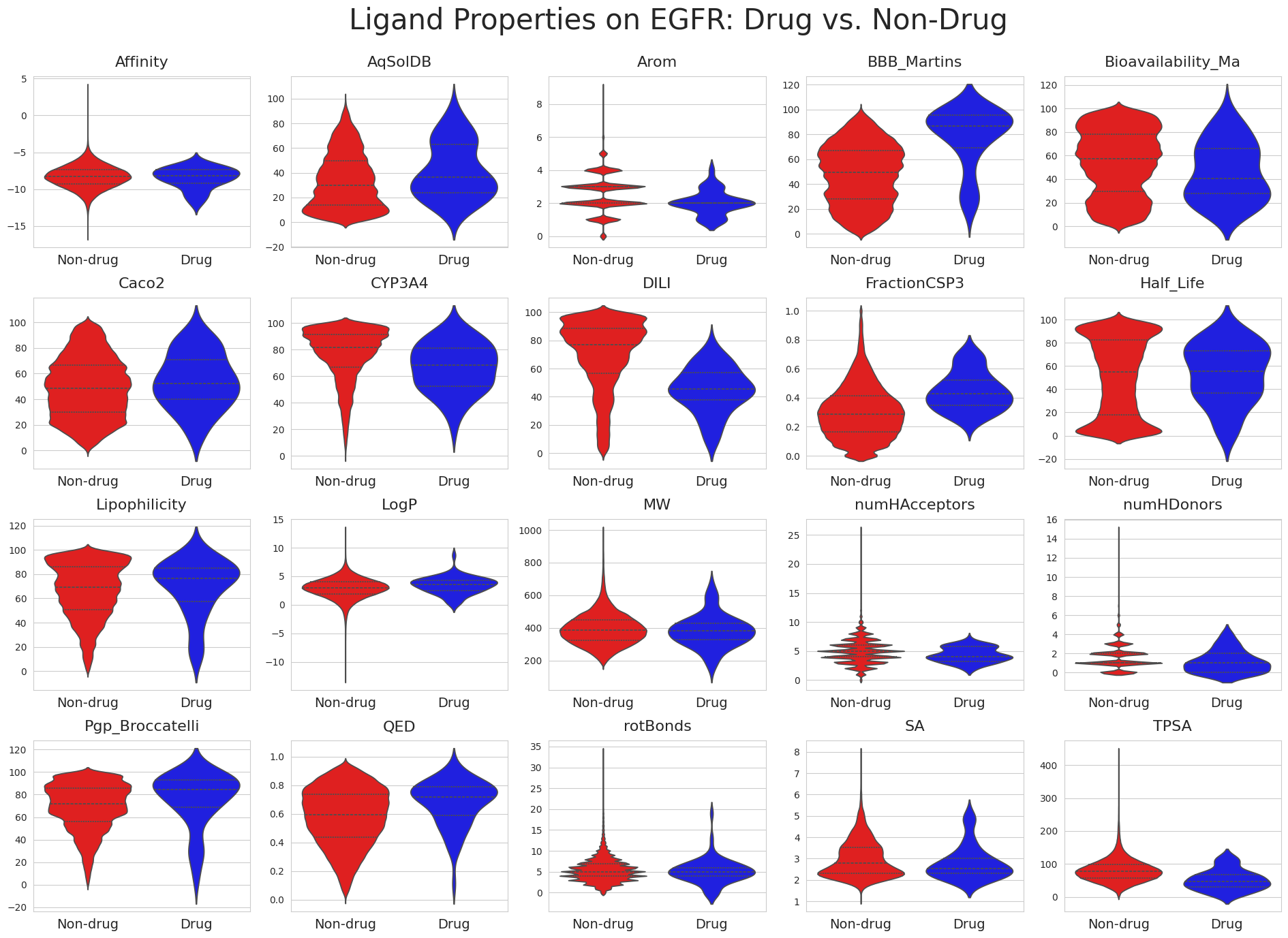}
\caption{Comparison of physicochemical and pharmacokinetic properties between known DRD2-targeting drugs (blue) and non-drug molecules (red) within 100,000-compound screening library. Violin plots illustrate key attributes such as affinity, molecular weight (MW), topological polar surface area (TPSA), blood-brain barrier permeability (BBB), and drug-induced liver injury (DILI), among others. The observed differences validate our objective selection, showing that drug-like molecules generally align with expected characteristics for CNS-active compounds, such as lower MW, optimized BBB permeability, and favorable toxicity profiles.}
\label{fig:drd2_lig}
\end{figure}

\begin{figure}[H]
    \centering
    \includegraphics[width=0.90\textwidth]{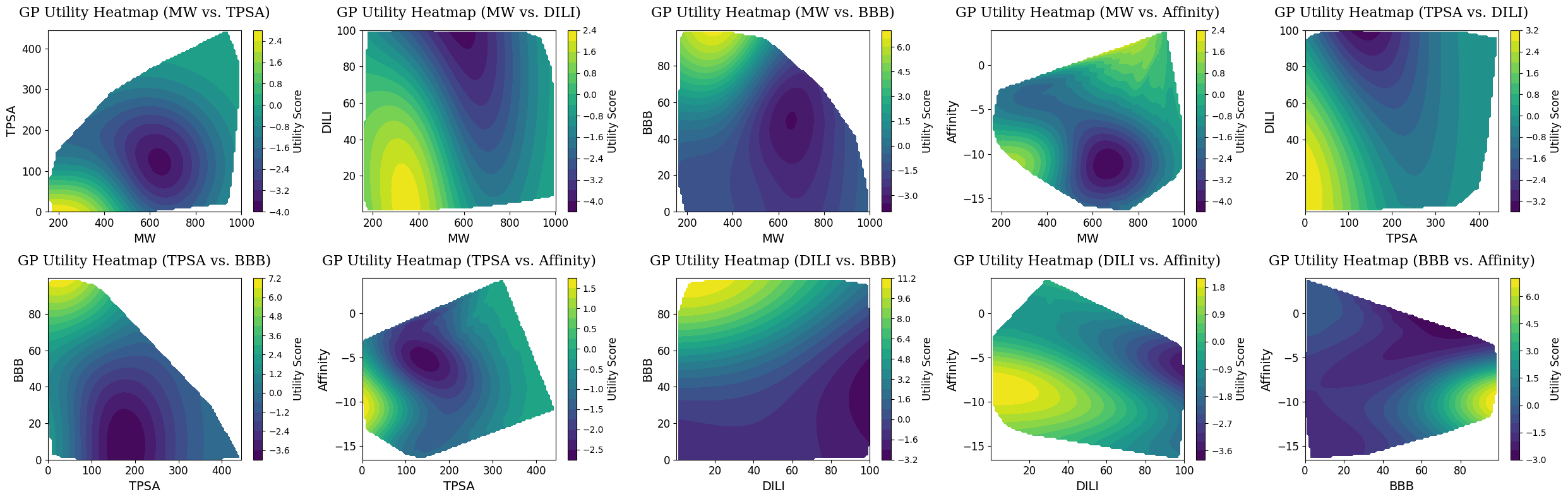}
    \caption{Predictive utility scores after BO on expert preference elicitation on DRD2. Heatmaps illustrate utility over two objectives while keeping others three at their mean. Results align well with established medicinal chemistry ranges, favoring optimal MW (200-400), TPSA (below 140), while maximizing BBB and minimizing DILI and binding affinity.}
    \label{fig:heatmap_drd2}
\end{figure}

\subsection{EGFR Experiments}
\begin{figure}[H]
\centering
\includegraphics[width=0.90\textwidth]{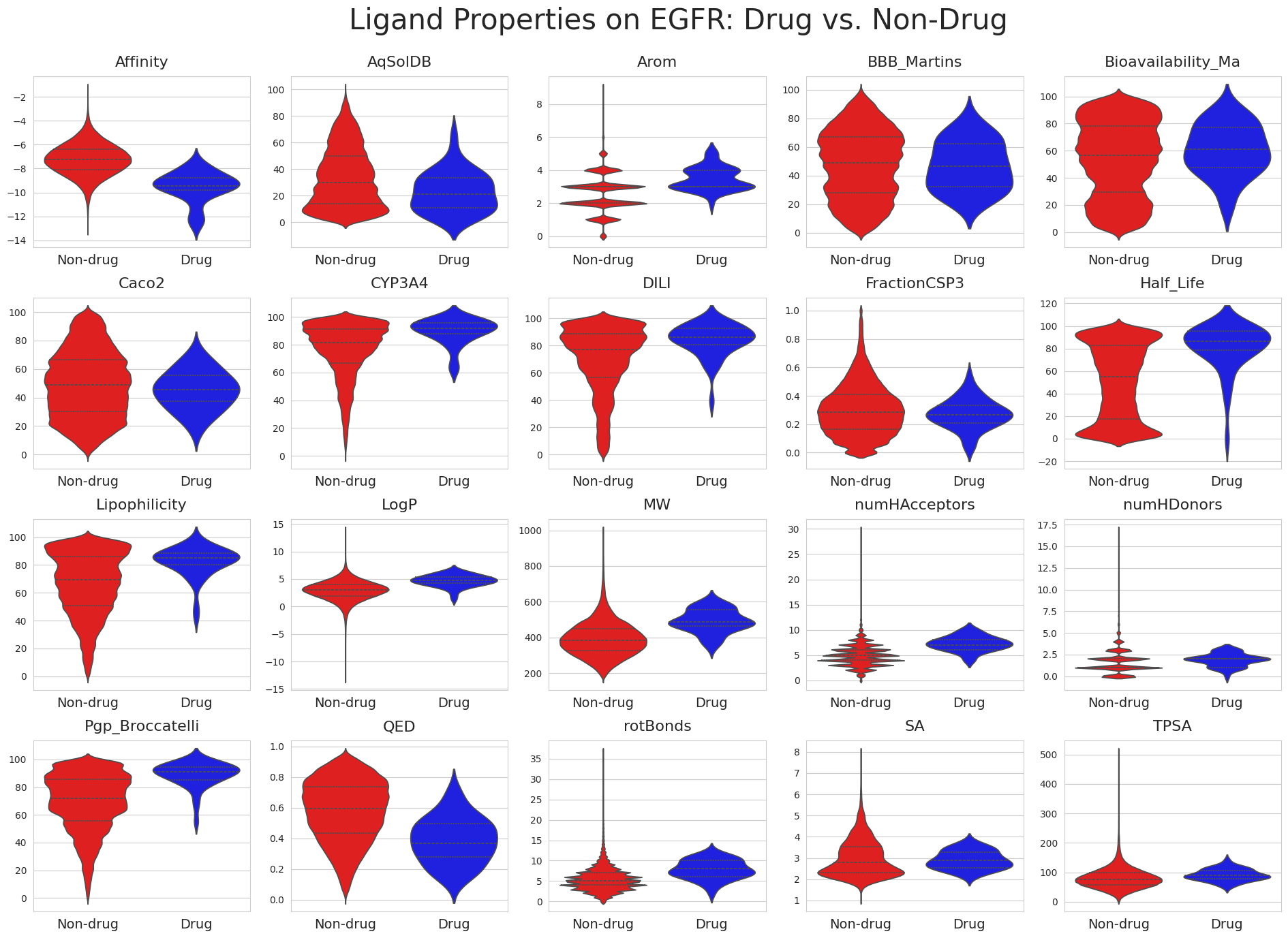}
\caption{Comparison of physicochemical and pharmacokinetic properties between known EGFR-targeting drugs (blue) and non-drug molecules (red) within the 100,000-compound screening library. The observed differences confirm that drug-like molecules generally exhibit characteristics favorable for kinase inhibition, including higher MW and optimized lipophilicity.}
\label{fig:egfr_lig}
\end{figure}

\begin{figure}[H]
    \centering
    \includegraphics[width=0.90\textwidth]{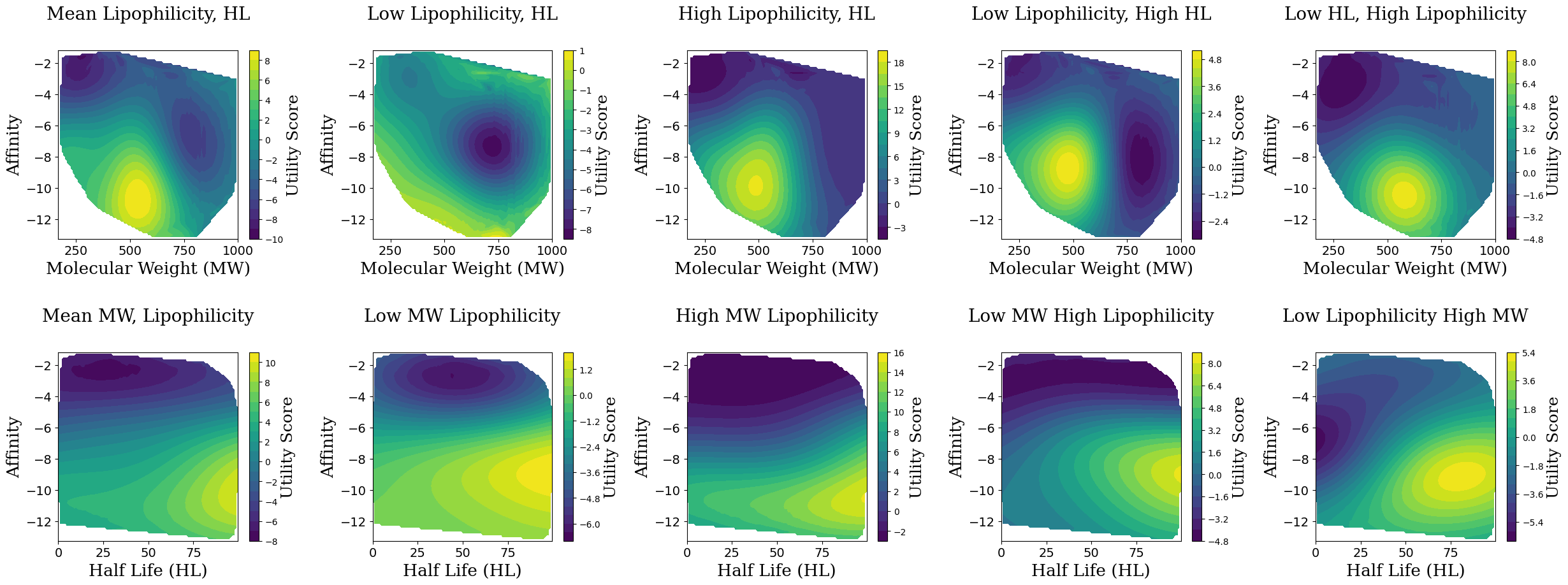}
    \caption{More on Gaussian process (GP) utility surfaces learned from expert preference data, illustrating the interplay among molecular weight (MW), half‐life (HL), affinity, and lipophilicity. Each heatmap shows the predicted utility (color scale) over two of these variables while holding the others fixed at the levels indicated in each title. Higher (yellow) regions correspond to more favorable trade‐offs according to the elicited expert preferences, providing insights for optimizing lead compounds in drug discovery.}
    \label{fig:more_elicitation}
\end{figure}

\subsection{Guidelines for Chemists}
\label{sec:guideline}
\begin{figure}[H]
\centering
\includegraphics[width=0.7\textwidth]{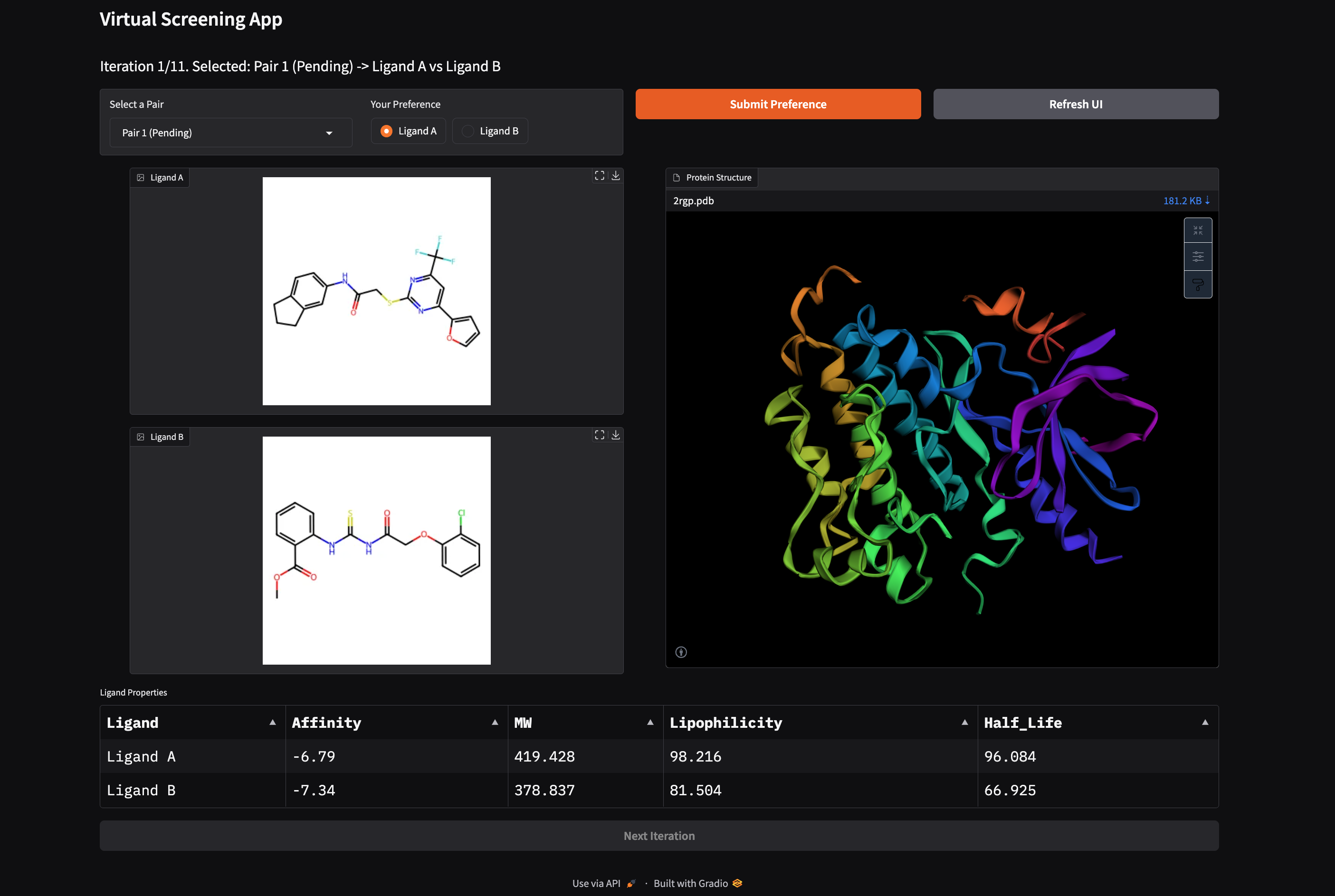}
\caption{Virtual Screening (VS) App built with Gradio for seamless interaction with chemists}
\label{fig:app_vs}
\end{figure}

The virtual screening app \ref{fig:app_vs} assists chemists in evaluating and comparing ligands by providing key molecular properties such as binding affinity, molecular weight (MW), lipophilicity, and half-life. It integrates SMILES-based molecular visualizations alongside numerical data, enabling users to analyze structural and chemical characteristics effectively. Chemists select their preferred ligand based on predefined criteria, and their selections contribute to refining the model’s predictive capabilities, improving its ability to identify promising drug-like candidates over time. However, the selection process is highly dependent on the biological target, as different proteins require distinct pharmacokinetic and pharmacodynamic considerations.

For example, targeting DRD2 in neuropharmacology necessitates prioritizing blood-brain barrier (BBB) permeability, as compounds must effectively penetrate the central nervous system while maintaining an appropriate balance between molecular weight and topological polar surface area (TPSA). Additionally, potential toxicity, such as predicted drug-induced liver injury (DILI), should be considered to ensure safety. In contrast, when designing inhibitors for EGFR in cancer therapy, selectivity, and affinity becomes paramount, as high target specificity minimizes off-target interactions and reduces systemic toxicity. A well-structured screening approach should reflect these protein-specific requirements, allowing chemists to weigh molecular properties appropriately when ranking compounds. Effective use of CheapVS requires some degree of expertise in medicinal chemistry, as misprioritizing criteria may lead to the selection of ineffective molecules.

\subsection{Synthetic Experiments}
\label{sec:synthetic}
\begin{figure}[H]
  \centering
  \includegraphics[width=0.77\textwidth]{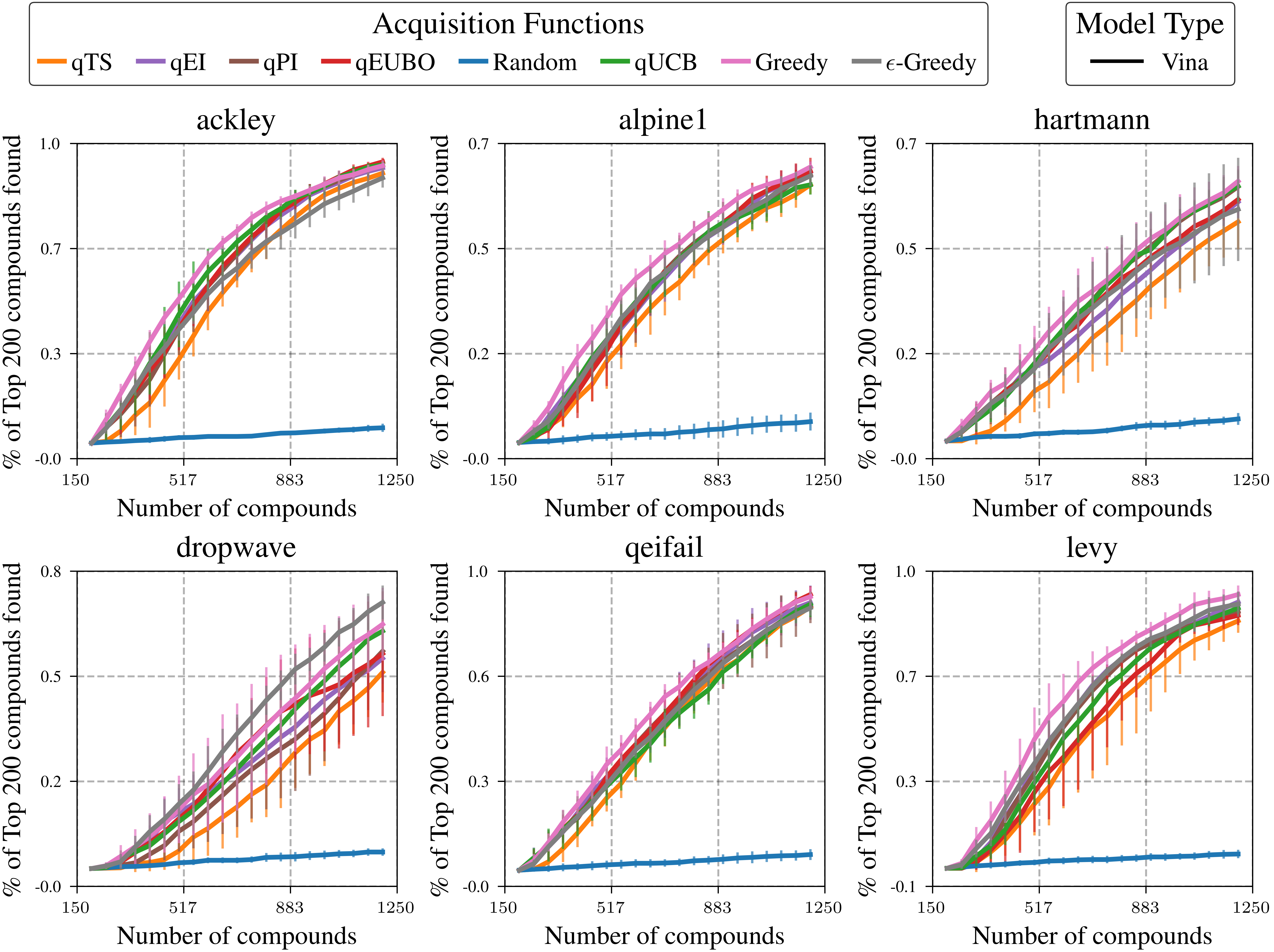}
  \caption{Preferential Multi-Objective Optimization results on multiple synthetic functions. The y-axis shows $\log(\textit{regret})$. The results compare multiple acquisition functions across various benchmark functions. Error bars indicate standard deviations across five seeds.}
  \label{fig:moavs_regret}
\end{figure}

\begin{figure}[H] %
\centering
\includegraphics[width=0.77\textwidth]{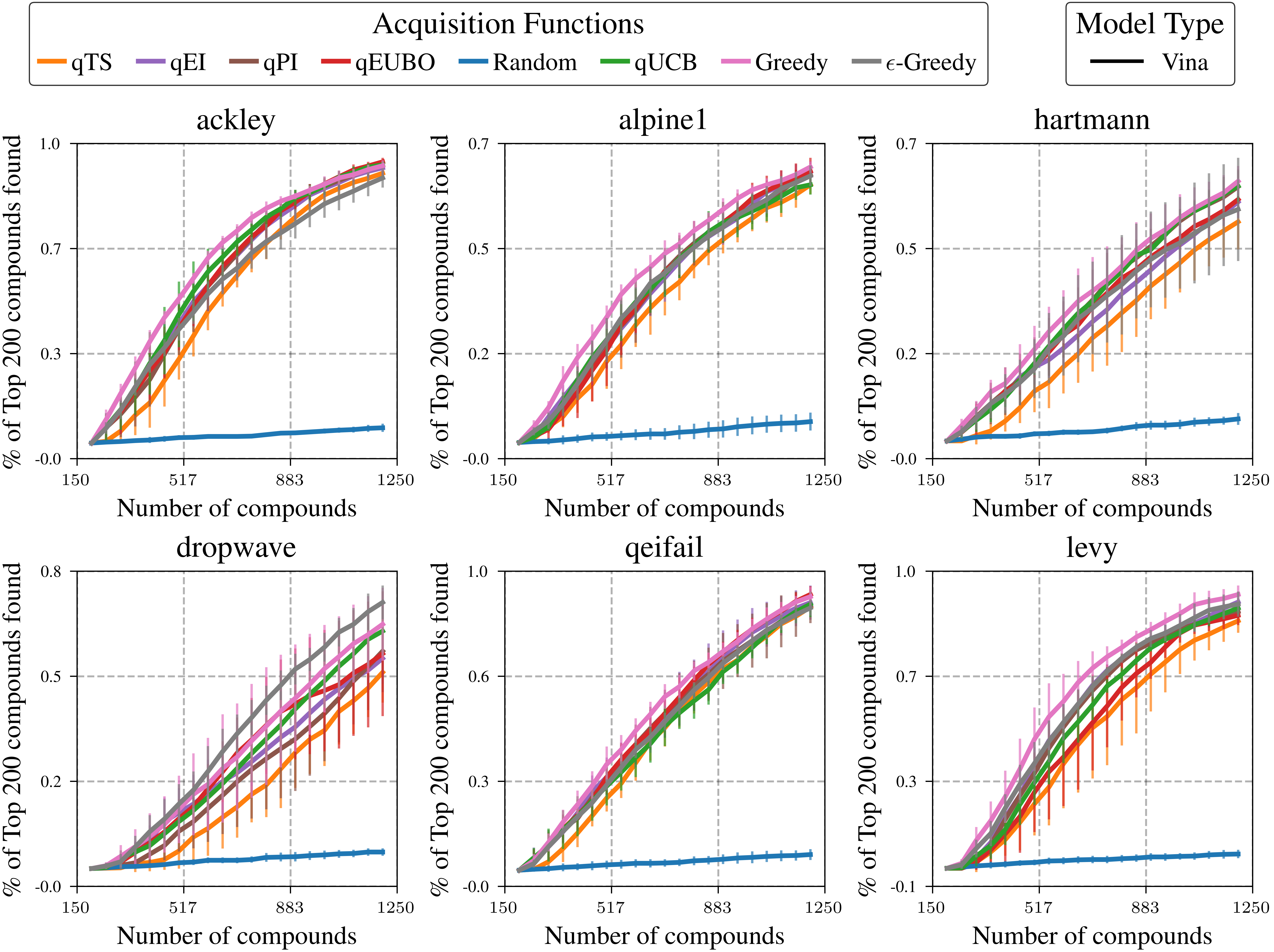}
\caption{Preferential Multi-Objective Optimization results on multiple synthetic functions. The y-axis shows accuracy. The results compare multiple acquisition functions across various benchmark functions. Error bars indicate standard deviations across five seeds.}
\label{fig:MOAVS_accuracy_graph}
\end{figure}

We examine how well \framework{} identifies high-utility solutions under various synthetic utility functions. Before running on real human preference data, we first test on synthetic functions. We create complex utility landscapes by modeling multi-dimensional molecular designs with benchmark functions: Ackley, Alpine1, Hartmann, Dropwave, Qeifail, and Levy. Each benchmark outputs a scalar ``utility,'' and we generate initial pairwise preference labels based on the corresponding utility values. In addition, we simulate four main objectives relevant to drug discovery: binding affinity, rotatable bonds, molecular weight, and LogP. For computational feasibility, we use a 20k-ligand subset sampled from the Dockstring library \cite{garcia2022dockstring}. Since the docking affinity values have already been computed for all compounds, we can determine both regret and accuracy. To ensure robustness, we repeat all experiments across five random seeds and report mean and standard deviation across runs. Furthermore, we evaluate a range of acquisition functions, including \texttt{qEUBO}, \texttt{qTS}, \texttt{qEI}, \texttt{qPI}, \texttt{qUCB}, \texttt{Greedy}, \texttt{$\epsilon$-Greedy}, and \texttt{Random}. Figure \ref{fig:moavs_regret} and Figure \ref{fig:MOAVS_accuracy_graph} displays the log regret and accuracy versus the number of compounds screened, illustrating the effectiveness of different acquisition strategies. The results show that regret consistently decreases and accuracy improves across all acquisition functions, with more advanced methods converging significantly faster than random baselines. These findings demonstrate that preferential BO effectively learns multi-objective trade-offs in synthetic benchmarks.

\section{Surrogate Model Performace}
\label{sec:surrogate}

\begin{table}[H]
\centering
\begin{tblr}{
  hline{1,5} = {-}{0.08em},
  hline{2} = {-}{},
}
Model Type                         & MSE Loss                            & NLPD                                                  \\
Fully-connected Neural Network         & 1.0568$\pm$0.0437                   & 1.4629$\pm$0.0198                                     \\
Decision Tree                      & 1.9785$\pm$0.2227                   & 1.7572$\pm$0.0548                                     \\
Gaussian Process (Tanimoto kernel) & \textbf{0.8549}$\pm$\textbf{0.0689} & \textbf{1.3389}\textbf{\textbf{$\pm$}}\textbf{0.0404} 
\end{tblr}
\caption{Comparison of model performance in predicting binding affinity values based on ligand fingerprints. The table reports the Mean Squared Error (MSE) Loss and Negative Log Predictive Density (NLPD) for different model types. Each model is trained on 6,000 samples using an 80/20 train/test split, and results are averaged over 20 random trials.}
\label{tab:surrogate1}
\end{table}

\begin{table}[H]
\centering
\begin{tblr}{
  hline{1,5} = {-}{0.08em},
  hline{2} = {-}{},
}
Model Type                 & Accuracy (\%)                       & ROC AUC                                     \\
Fully-connected Neural Net & 0.9505$\pm$0.0146                   & \textbf{0.9913}$\pm$\textbf{0.0081}                            \\
Decision Tree              & 0.7853$\pm$0.0285                   & 0.7858$\pm$0.029                             \\
Pairwise Gaussian Process  & \textbf{0.9563}$\pm$\textbf{0.0146} & 0.9724$\pm$0.0161
\end{tblr}
\caption{Comparison of utility model performance in predicting preference-based rankings from ligand properties on Ackley function. The table reports the classification accuracy and ROC-AUC of different model types. Each model is trained on 1,000 samples using an 80/20 train/test split, and results are averaged over 20 random trials. The Pairwise Gaussian Process achieves the highest classification accuracy and second highest ROC-AUC, demonstrating superior performance in modeling pairwise preferences and learning utility functions from ligand physicochemical properties.}
\label{tab:surrogate2}
\end{table}

\section{Diffusion Model Training: Hyperparameters and Performance Results}

\begin{table}[H]
\centering
\begin{tblr}{
  hline{1,9} = {-}{0.08em},
  hline{2} = {-}{},
  colspec={lcccc},
}
Model~                      & DockScan22                           & EDM-S (Pre-train)                    & EDM-S~~(Fine-tune)                 & EDM-S (EGFR)                       \\
Parameters initialized from & Random                               & Random                               & EDM-S (Pre-train)                  & EDM-S~~(Fine-tune)                 \\
Batch Size                  & 256                                  & 256                                  & 256                                & 64                                 \\
Number of Epochs            & 150                                  & 2.32                                 & 140                                & 640                                \\
Dataset train on            & PDBScan22                            & 11M synthetic data                   & PDBScan22                          & ChemDiv 10k                        \\
Learning Rate               & $1.8 \times 10^{-3}$ & $1.8 \times 10^{-3}$ & $1 \times 10^{-3}$ & $2 \times 10^{-3}$ \\
Diffusion steps             & 20                                   & 20                                   & 20                                 & 10                                 \\
$\sigma_{\text{data}}$      & 32                                   & 32                                   & 32                                 & 5                                  
\end{tblr}
\caption{Hyperparameters for training EDM-S and DockScan22}
\end{table}

\begin{table}[htb!]
\centering
\setlength{\tabcolsep}{4pt} %
\begin{tabular}{lccccccc}
\toprule
\multicolumn{1}{c}{} &
  \multicolumn{2}{c}{\begin{tabular}[c]{@{}c@{}}PoseBuster V1\\ Top-1 RMSD (Å)\end{tabular}} &
  \multicolumn{2}{c}{\begin{tabular}[c]{@{}c@{}}PoseBuster V2\\ Top-1 RMSD (Å)\end{tabular}} &
  \multicolumn{2}{c}{\begin{tabular}[c]{@{}c@{}}PDBBind\\ Top-1 RMSD (Å)\end{tabular}} &
  \begin{tabular}[c]{@{}c@{}}Inference time \\ on 1 A100\end{tabular} \\
\multicolumn{1}{c}{Metrics}        & \% $<$ 2Å      & \% $<$ 5Å   & \% $<$ 2Å     & \% $<$ 5Å     & \% $<$ 2Å     & \% $<$ 5Å   & seconds     \\ \hline
DIFFDOCK-S (40)                    & 24             & 45.1        & -   & -   & 31.1          & - & \textbf{10} \\
DIFFDOCK (40)                      & 37.9           & 49.3        & -   & -   & \textbf{38.2} & \textbf{62} & 30          \\ \hline 
AlphaFold 3 (25)                   & 76.4           & - & \textbf{80.5} & -   & -   & - & 340         \\
Chai-1 (25)                        & \textbf{77.05} & - & -   & -   & -   & - & 340         \\ \hline \hline
\textbf{DockScan22 (40)} & 54.1           & 77.8        & 58.8          & 81.4          & 34.1          & 56          & \textbf{10} \\
\textbf{EDM-S (40)}                & 30             & \textbf{91} & 32.2          & \textbf{92.1} & -   & - & \textbf{10} \\
\bottomrule
\end{tabular}
\caption{Performance comparison on PDBBind and PoseBuster benchmarks, with models sampling 40 or 25 ligand poses per protein-ligand pair. Highlighted rows show our proposed methods, offering competitive accuracy with significantly lower runtime.}
\label{tab:training_result}
\end{table}

\begin{figure}[htbp]
\centering
\includegraphics[width=1.0\textwidth]{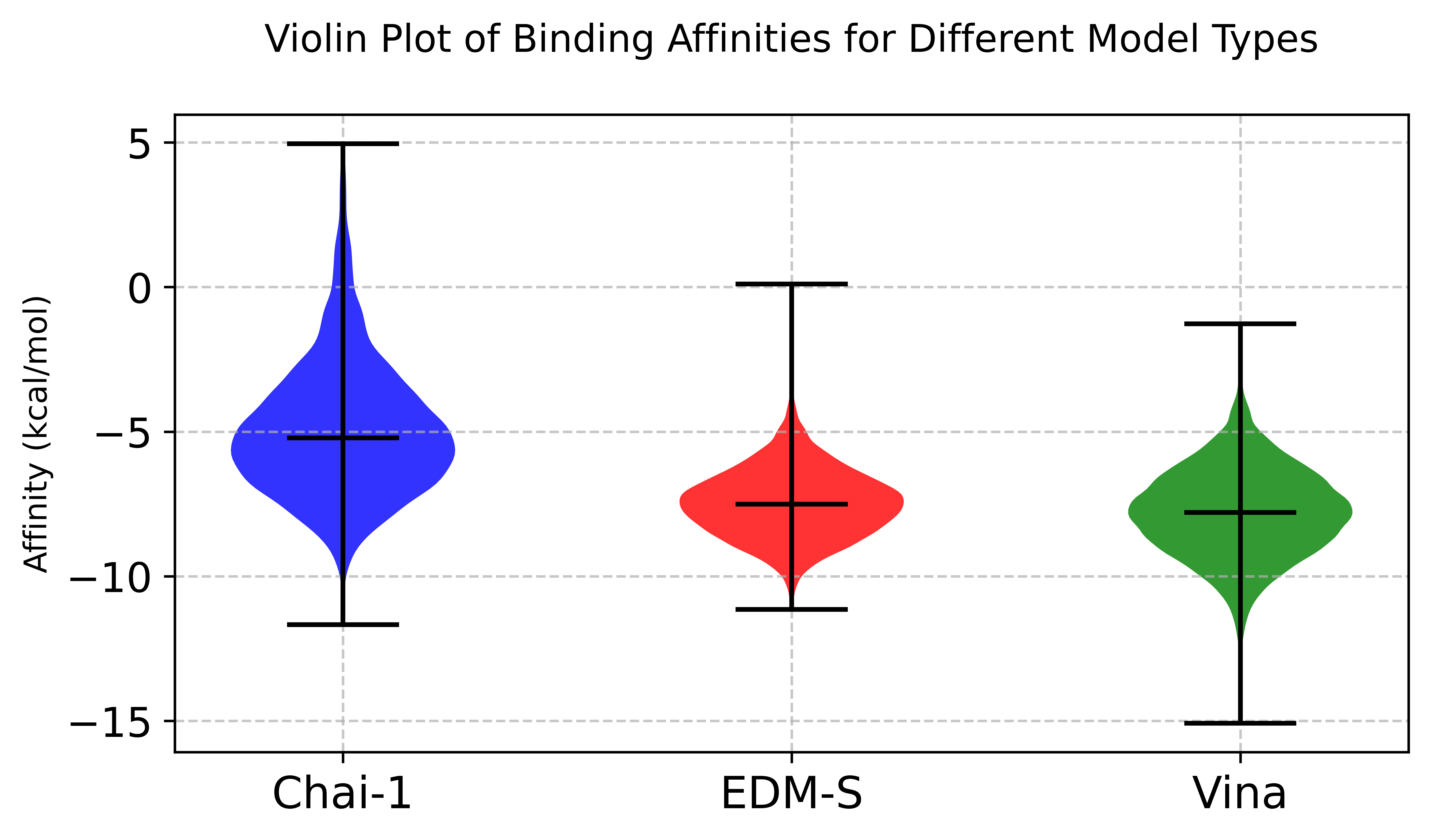}
\caption{Violin plot of binding affinities (kcal/mol) for different docking models on the EGFR protein with 6000 ligands. Vina achieves the lowest median binding affinity, followed by EDM-S, while Chai exhibits the weakest binding.}
\label{fig:box_affinity}
\end{figure}

\begin{figure}[H]
    \centering
    \includegraphics[width=1.0\textwidth]{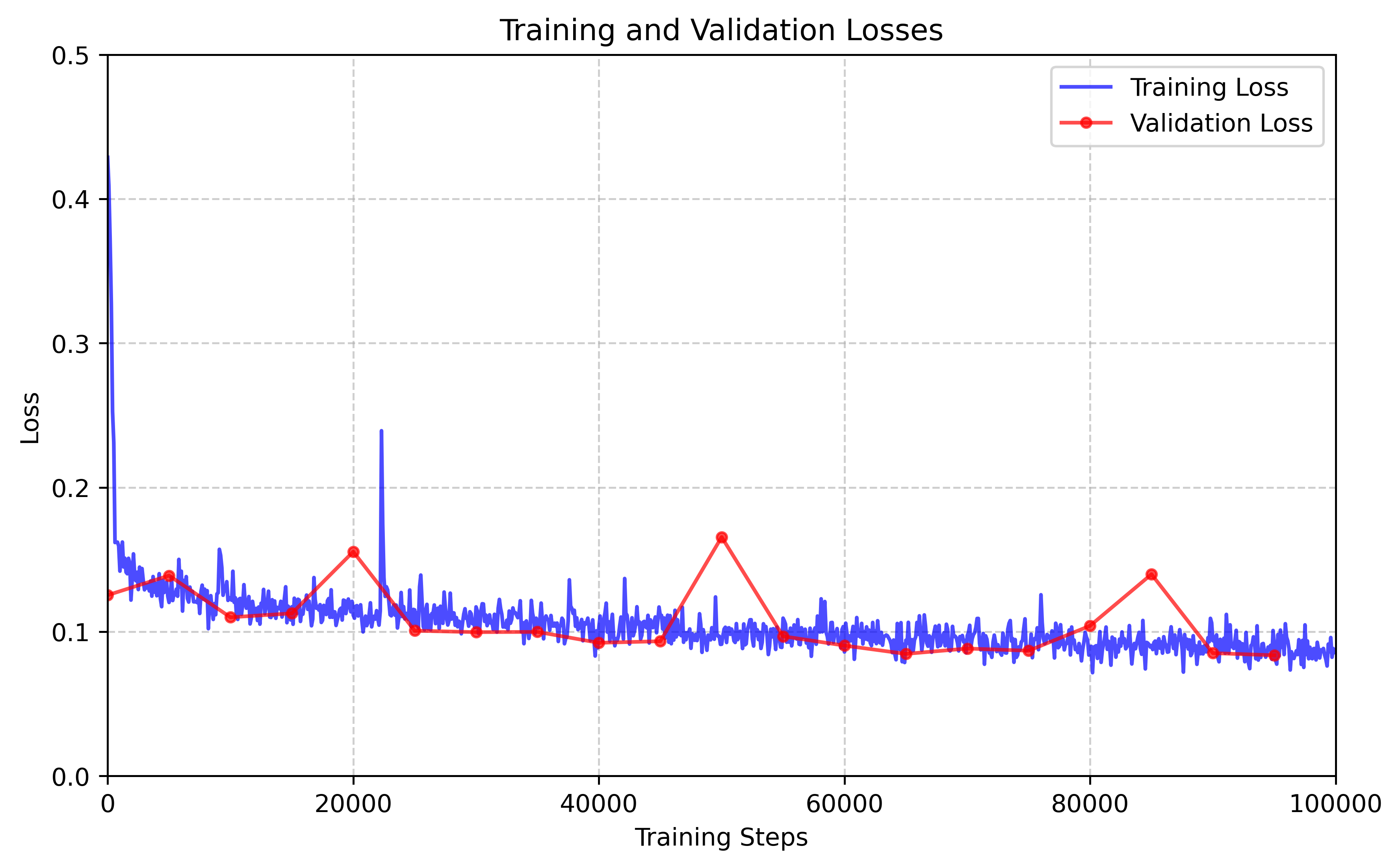}
    \caption{Training and validation loss of EDM-S on EGFR protein with 10k ligands.}
    \label{fig:edm_s_loss_2rgp}
\end{figure}

\end{document}